\documentclass[10pt,twocolumn,letterpaper]{article}

\usepackage[pagenumbers]{cvpr} 
\usepackage{nicematrix}    

\usepackage[accsupp]{axessibility} 
\usepackage{tcolorbox}
\definecolor{cvprblue}{rgb}{0.21,0.49,0.74}
\usepackage[pagebackref,breaklinks,colorlinks,allcolors=cvprblue]{hyperref}
\def\paperID{} 
\def\confName{CVPR}
\def\confYear{2026}

\title{See Further, Think Deeper: Advancing VLM's Reasoning Ability with Low-level Visual Cues and Reflection}

\author{
Zhiheng Wu$^{1}$\thanks{Equal contribution.}$^{*}$, Tong Wang\thanks{Independent researcher.} $^{*}$, Shuning Wang$^{2}$, Naiming Liu$^{3}$, Yumeng Zhang$^{1}$\thanks{Corresponding author.} \\
$^{1}$Baidu Inc.,  $^{2}$Zhejiang University, $^{3}$Harbin Institute of Technology
}

\begin{document}
\maketitle
\begin{abstract}
   Recent advances in Vision-Language Models (VLMs) have benefited from Reinforcement Learning (RL) for enhanced reasoning. However, existing methods still face critical limitations, including the lack of low-level visual information and effective visual feedback. To address these problems, this paper proposes a unified multimodal interleaved reasoning framework \textbf{ForeSight}, which enables VLMs to \textbf{See Further} with low-level visual cues and \textbf{Think Deeper} with effective visual feedback. First, it introduces a set of low-level visual tools to integrate essential visual information into the reasoning chain, mitigating the neglect of fine-grained visual features. Second, a mask-based visual feedback mechanism is elaborated to incorporate visual reflection into the thinking process, enabling the model to dynamically re-examine and update its answers. Driven by RL, ForeSight learns to autonomously decide on tool invocation and answer verification, with the final answer accuracy as the reward signal. To evaluate the performance of the proposed framework, we construct a new dataset, Character and Grounding SalBench (CG-SalBench), based on the SalBench dataset. Experimental results demonstrate that the ForeSight-7B model significantly outperforms other models with the same parameter scale, and even surpasses the current SOTA closed-source models on certain metrics.
\end{abstract}    
\section{Introduction}


\begin{figure}[htbp]
  \centering 
  \includegraphics[width=1\linewidth]{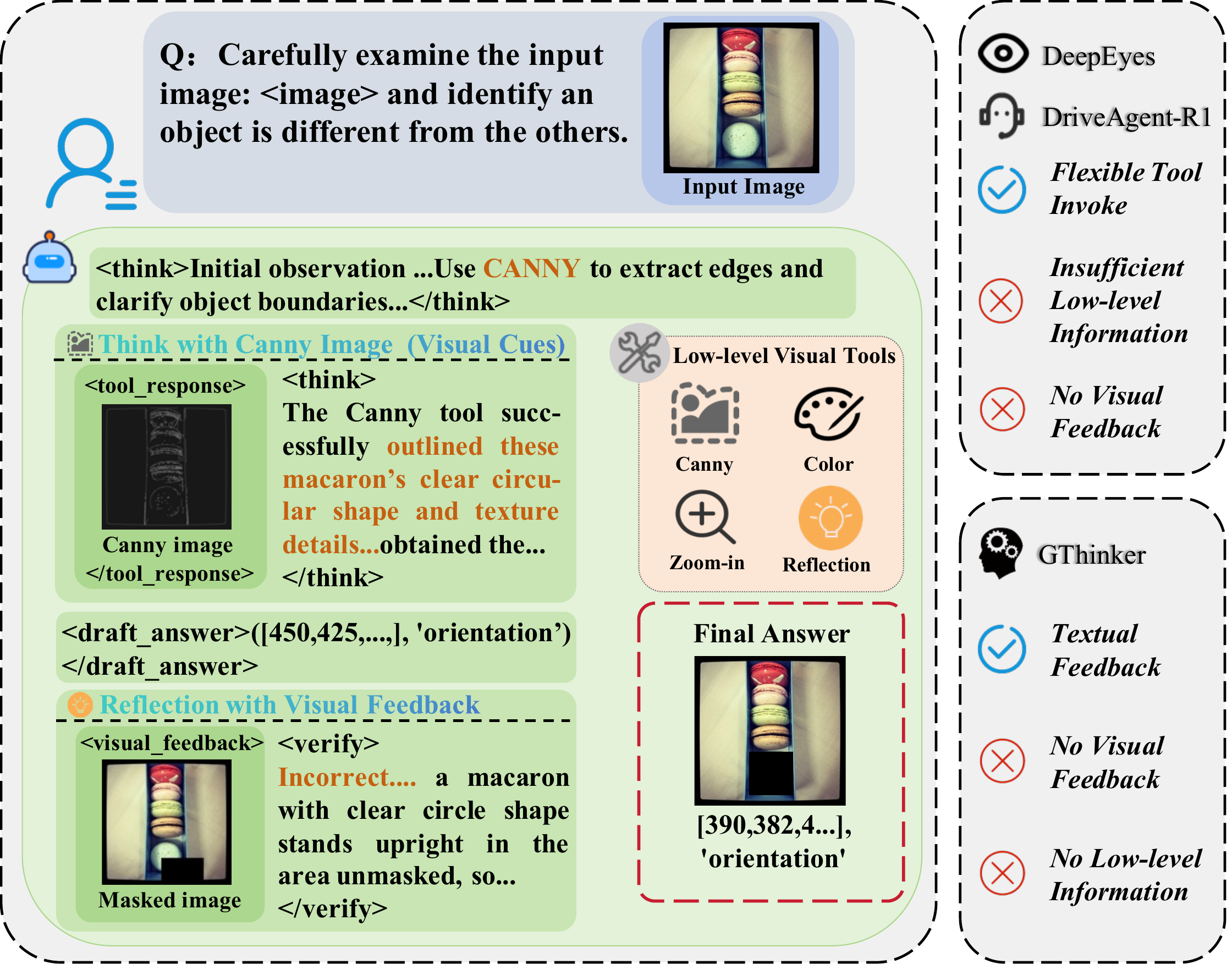}
  \caption{\textbf{An example of ForeSight's inference process.} Our model dynamically invokes relevant tools during reasoning and integrates their outputs for deeper inference. After generating an initial draft answer, the model produces visual feedback based on this answer and conducts further reflection to verify the correctness of the answer. } 
  \label{fig:introduction} 
\end{figure}

\noindent
Recent advances in Large Language Models (LLMs) have shifted the research focus from mere language understanding toward the exploration and modeling of complex reasoning capabilities~\cite{he-etal-2025-breaking,zheng2025knowledge,chiang2023vicuna,peng2023instruction}.
With the rapid progress of Reinforcement Learning (RL), LLMs have demonstrated remarkable abilities in tool-augmented and reflective reasoning, leading to substantial improvements in complex reasoning and decision-making tasks~\cite{wei2025autotirautonomoustoolsintegrated,das2024mathsenseitoolaugmentedlargelanguage,feng2025retoolreinforcementlearningstrategic}. Furthermore, RL has also been adapted to Vision-Language Models (VLMs) to enhance the ability of solving complex tasks~\cite{huang2025vision,shen2025vlm,peng2025lmm}.

While the integration of RL has significantly enhanced VLMs' reasoning ability, their reasoning processes are largely confined to the language modality~\cite{team2025kimi,guo2025seed1}. The text-only reasoning for visual tasks contradicts the process by which humans perform visual reasoning tasks. A sound reasoning process should involve both the visual and language modalities. When the acquired visual information is insufficient to support the reasoning process, humans will actively acquire additional visual information. For VLMs, such text-only reasoning may lead to the 
neglect of fine-grained visual features. This issue becomes particularly severe when the reasoning process lengthens.
Hence, recent studies have introduced the interleaved Multimodal Chain-of-Thought (i-MCoT)~\cite{gao2025interleaved} paradigm to support step-by-step cross-modal reasoning. For example, DeepEyes~\cite{zheng2025deepeyes} equips the models with the capability of ``thinking with images" through end-to-end RL. DriveAgent-R1~\cite{zheng2025driveagent} introduces a Hybrid-Thinking framework that adaptively switches between efficient text-based and in-depth tool-based reasoning. 

Nevertheless, existing i-MCoT approaches still present two main limitations: \textbf{(1) the lack of sufficient low-level visual information}; \textbf{(2) the absence of an effective visual feedback mechanism.} For the former problem, SalBench~\cite{huynh2025vision} uncovered a critical issue: state-of-the-art large VLMs perform poorly in detecting salient and intuitive visual anomalies in images. This underperformance is particularly notable because these low-level perceptual tasks may seem quite simple to humans. And VAT~\cite{liu2025VAT} demonstrated that introducing low-level visual information, such as edge contours, binarized maps, or semantic sketches, can significantly enhance the performance of VLMs on visual tasks.
Although both DeepEyes and DriveAgent-R1 have integrated image tools into their reasoning processes, their tools primarily focus on RoI region scaling and image understanding tasks such as 3D object detection and depth map generation, while overlooking the utilization of low-level visual information. To address this limitation, we provide a set of fundamental low-level visual tools and employ RL to enable the model to autonomously decide whether and when to invoke these tools during inference, thereby achieving interleaved vision-language reasoning. This mechanism endows the model with the ability to proactively scrutinize visual details when necessary.

For the latter problem, most existing VLMs lack effective reflection with visual feedback signals in their reasoning processes~\cite{liu2023visual, zhu2023minigpt, dai2023instructblip}. The process from reasoning to generating answers is open-loop: the final answer is derived through reasoning based on existing visual or textual information without reflection. And once an answer is produced, the model will not make any adjustments to it. Such a one-way reasoning process typically only achieves limited intelligence. We aim to incorporate a visual feedback mechanism into the reasoning chain, transforming the original one-way ``reasoning → answer" process into a closed-loop cycle of ``reasoning → answer → visual feedback → reasoning → answer...". Through this visual feedback, the model can re-examine its own answers and thereby determine whether further modifications to the answers are necessary. 
Specifically, we design a mask-based visual feedback mechanism. We utilize the bounding boxes of the VLM’s initial answer to mask the corresponding regions of the original image, then feed the masked image back into the VLM to prompt it to reflect on the initially predicted answer and decide whether to generate an improved answer. 
It is worth noticing that GThinker~\cite{zhan2025gthinker} also introduces a reflection mechanism guided by visual cues, but its reasoning process remains text-centric and does not achieve truly interleaved multimodal reasoning.

In this paper, we propose a unified multimodal interleaved reasoning framework \textbf{ForeSight}, which possesses two key characteristics. The first one is that we introduce a series of low-level visual tools to incorporate essential low-level visual information into the reasoning process, enabling VLMs to \textbf{See Further} during inference. The second one is the integration of reflection with a mask-based visual feedback mechanism. Such a mechanism grants VLMs the ability to \textbf{Think Deeper} and allows the model to re-examine its generated answers and update them when necessary. The entire framework is driven by RL, with the accuracy of the final answer serving as the reward signal. As shown in Fig.~\ref{fig:introduction}, the trained VLM can autonomously perform tool invocation or reflection. To validate the effectiveness of the proposed framework, we constructed a new dataset \textbf{C}haracter and \textbf{G}rounding SalBench (CG\text{-}SalBench) based on the SalBench~\cite{huynh2025vision} dataset and conducted experiments on this newly built dataset. Experimental results show that compared with Qwen2.5-VL-7B, ForeSight-7B achieves significant performance improvements, with the IoU increasing by 29.68\% and the F1 score rising by 19.27\%.

Our main contributions are summarized as follows:

\begin{itemize}
\item We are the first to introduce reflection with visual feedback into the reasoning process of VLMs, upgrading the open-loop reasoning process to a closed-loop one.
\item We propose an RL-driven ForeSight framework that integrates low-level visual information and reflection with visual feedback into the reasoning process. 
\item We present a new dataset, CG\text{-}SalBench, built upon SalBench. Experimental results show that our method outperforms the baseline counterpart by a large margin. 
\end{itemize}

\label{sec:intro}

\section{Related work}

\noindent
\textbf{Vision-Language Model.}
Recent advances in VLMs mark a shift from modular pipelines to end-to-end architectures that enable seamless multimodal integration. Early frameworks such as BLIP-2~\cite{li2023blip} and LLaVA~\cite{liu2023visual,liu2024llava, guo2024llava,lin2023video} achieve cross-modal alignment through learnable adapters or cross-attention modules, effectively bridging visual and linguistic representations. Building on this paradigm, open-source models like Qwen-VL~\cite{bai2023qwenvlversatilevisionlanguagemodel, wang2024qwen2, bai2023qwen}, MiniCPM-V~\cite{yao2024minicpmvgpt4vlevelmllm,yu2025minicpm}, and InternVL~\cite{chen2024internvl, gao2024mini, wang2025internvl3} further enhance fine-grained perception and multi-image reasoning, driving progress in unified multimodal learning.
More recently, studies such as GThinker~\cite{zhan2025gthinker}, Chain-of-Action~\cite{pan2024chain}, and AgentThink~\cite{qian2025agentthink} have extended VLMs toward reasoning and tool-augmented paradigms, aiming to model higher-level cognitive processes. However, most current approaches still rely on internal knowledge without leveraging external visual tools, which limits their capacity for complex visual reasoning and real-world interaction.

\noindent
\textbf{Multimodal Chain-of-Thought Reasoning.}
Chain-of-Thought (CoT)~\cite{li2024towards,wei2022chain} has significantly enhanced the complex problem-solving capabilities of LLMs by guiding them through step-by-step reasoning. This paradigm has been successfully extended to VLMs, giving rise to the research framework of Multimodal Chain-of-Thought (MCoT). Early MCoT methods adopted a ``perceive-then-reason” paradigm, but decoupling vision and language often led to loss of critical visual details. Recent works like LLaVA-CoT ~\cite{xu2025llava}, Virgo ~\cite{du2025virgo}, and Mulberry ~\cite{yao2024mulberry} introduced ``Slow Thinking” mechanisms to improve systematic reasoning, yet they are commonly task-specific and generalize poorly in open-domain settings. GThinker ~\cite{zhan2025gthinker} proposed a more flexible, prompt-driven approach with vision-guided reflection to improve cross-task generalization and interpretability. Recent MCoT~\cite{jiang2025vlm,su2025openthinkimg,wang2025pixel} methods have begun to explore tool-assisted visual reasoning. However, most MCoT~\cite{wang2025multimodal}  methods are still limited by their reliance on the model's internal knowledge or passive processing of visual inputs, as they currently lack a general-purpose, model-autonomous tool-calling mechanism.

\noindent
\textbf{Reinforcement Learning for VLM.}
In recent years, RL has gradually emerged as a key approach for enhancing the reasoning capabilities and behavioral alignment of VLMs. In contrast to early supervised methods, such as LLaVA-Reasoner~\cite{zhang2024improve} and MPO~\cite{singla2024dynamic}, that relied on human-annotated reasoning trajectories or preference data, recent studies have increasingly adopted outcome-driven reward mechanisms, enabling models to autonomously explore effective strategies through trial and error. For instance, DeepSeek-R1~\cite{guo2025deepseek} demonstrated that using only the correctness of the final answer as a reward signal is sufficient to drive the emergence of self-verification and reflective reasoning capabilities. This paradigm has been preliminarily adapted to multimodal settings to refine reasoning in structured tasks such as mathematics and science. However, general visual reasoning 
commonly involves ambiguous perception, open-ended objectives, and dynamic decision-making, making it difficult to rely on a single, deterministic reward criterion.


\begin{figure}[htbp]
  \centering 
  \includegraphics[width=1\linewidth]{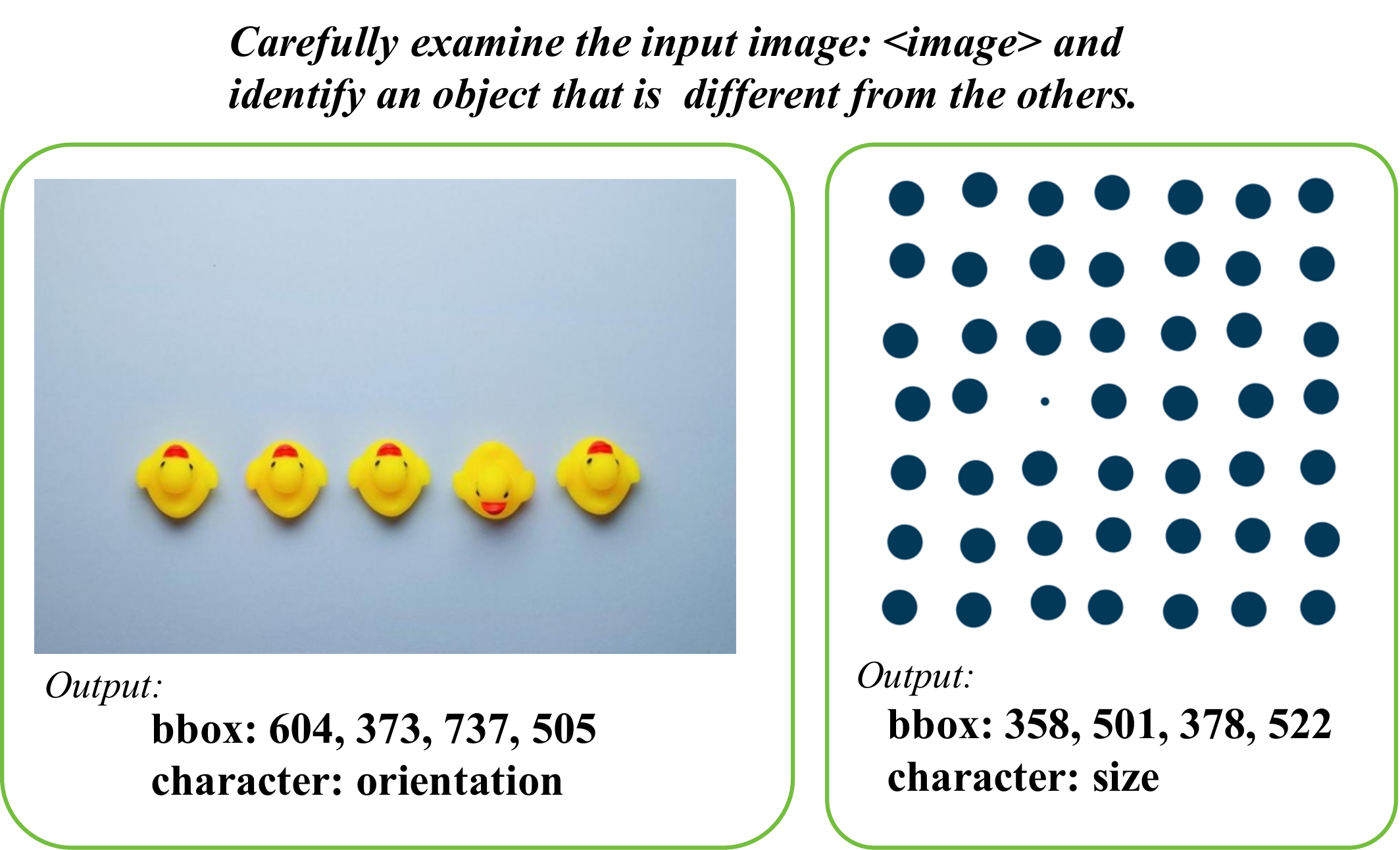}
  \caption{\textbf{The illustration of CG-SalBench.} Each sample contains multiple similar distractors and one distinct target, which differs in one or more specific visual characters. The task is to output the bounding box (bbox) and character of a specific object. The left image is a natural image (O3), and the right image is a synthetic image (P3).} 
  \label{fig:bench} 
\end{figure}

\begin{figure*}[htbp]
  \centering 
  \includegraphics[width=1\linewidth]{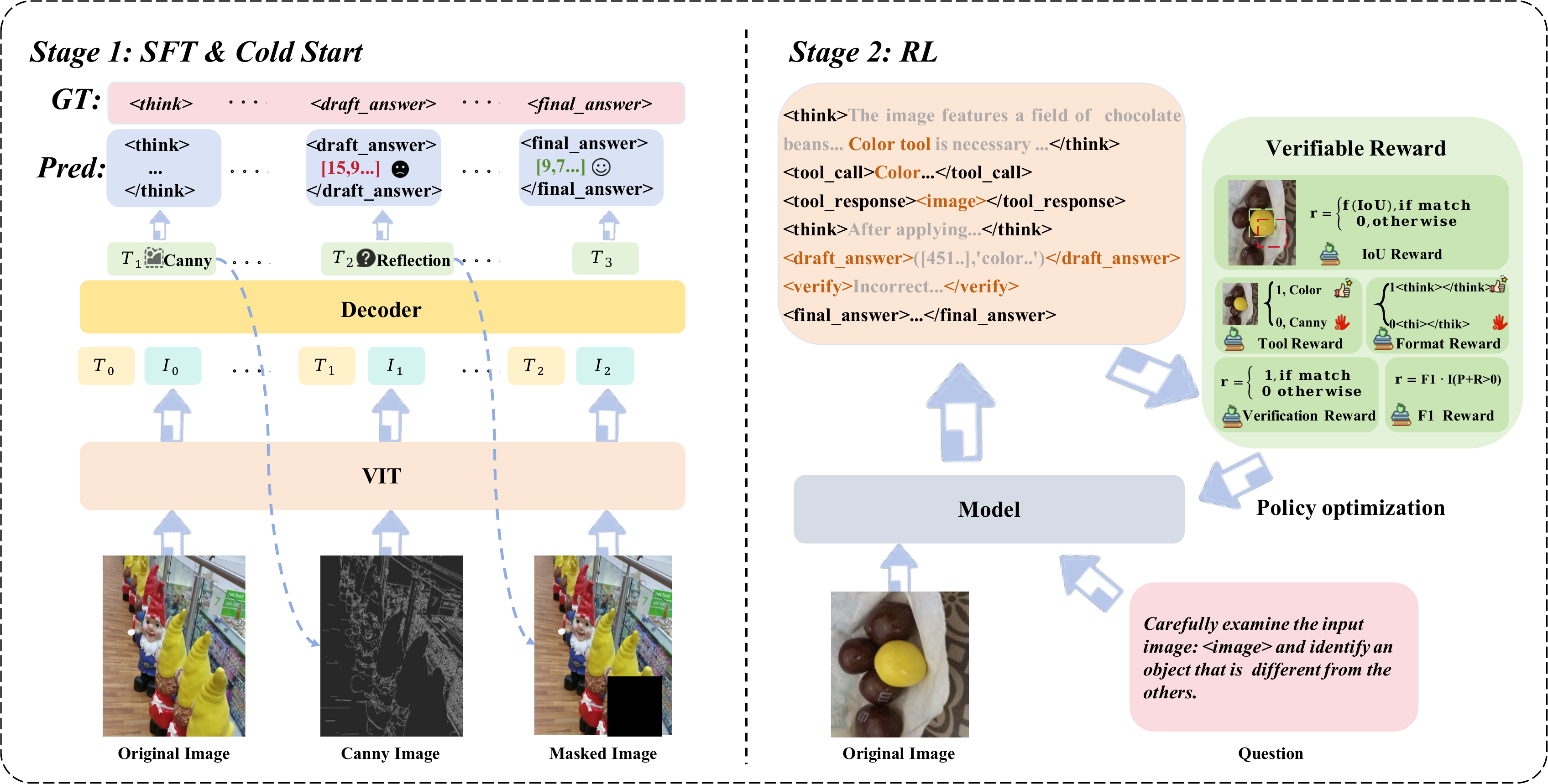}
  \caption{\textbf{Overview of ForeSight’s two-stage framework.} Stage 1: We use high-quality tool calls and visual reflection data for supervised fine-tuning to teach the model how to utilize external tools and generate coherent inference trajectories effectively. Stage 2: We further apply reinforcement learning and the GRPO algorithm, combined with tool-based rewards and several other complementary rewards, to achieve end-to-end optimization.} 
  \label{fig:framework} 

\end{figure*}

\section{Method}

In this section, we first introduce CG-SalBench in Section \ref{method:benchmark}, then outline our ForeSight framework in Section \ref{method:ForeSight}. Next, we elaborate on the design of the visual tools and reflection mechanisms in Section \ref{method:tool}, describe the construction of the cold start dataset in detail in Section \ref{method:Training Data}, and finally elaborate on the RL setup in Section \ref{method:RL}.

\subsection{CG-SalBench}\label{method:benchmark}

We introduce Odd-One-Out \textbf{C}haracter and \textbf{G}rounding Saliency Benchmark, \textit{i.e.}, CG-SalBench, a novel extension of SalBench~\cite{huynh2025vision}. SalBench is a benchmark designed to evaluate the low-level visual saliency perception capabilities of large vision-language models. Its core task is to identify the distinctive visual character of Odd-One-Out objects. Specifically, each image consists of multiple highly similar distractors and one target object that differs significantly along one or more dimensions—such as color, orientation, size, shape, focus, position, or pattern. This benchmark includes both the controlled synthetic dataset P3 (2589 images) and the complex, diverse real-world dataset O3 (2001 images)~\cite{kotseruba2020saliency}, thereby assessing whether models possess fundamental perceptual abilities aligned with human visual attention mechanisms. As shown in Fig.~\ref{fig:bench}, CG-SalBench requires models to not only identify the anomalous visual character of the Odd-One-Out object, but also to provide its precise bounding box. It aims to push the boundaries of saliency detection. Based on the CG-SalBench, this study partitioned the total sample set into a training set (3929 samples) and a test set (661 samples). The test set consists of 252 natural image samples and 409 synthetic samples.

\subsection{ForeSight}\label{method:ForeSight}

We propose ForeSight, a unified framework for multimodal tool-augmented reasoning~\cite{qian2025toolrl}, which enables the model to proactively invoke a set of low-level visual tools during the i-MCoT process. 
By incorporating these visual operations, the model can more effectively capture fine-grained and fundamental visual information, leading to better performance on ``what-you-see-is-what-you-get" tasks. This capability is realized through the integration of the inherent linguistic competence of LLMs and the decision-making ability learned via Group Relative Policy Optimization (GRPO)~\cite{guo2025deepseek}.

As illustrated in Fig.~\ref{fig:framework}, given a question $Q$ and an image input $I$, the model is required to autonomously decide at each step of the textual CoT reasoning whether sufficient information is available to produce a correct answer. If the model determines that an accurate answer can be derived, it directly outputs the answer. Otherwise, it invokes appropriate low-level visual tools to process and enhance the image, and then re-engages in reasoning based on the updated visual input. This iterative procedure continues until the model believes a correct answer has been reached or the maximum number of tool invocations is exceeded. In addition, we incorporate a visual reflection mechanism into our framework to enhance the model’s capability for self-evaluation and verification during multimodal reasoning. Specifically, after generating an initial draft answer, the model invokes a visual reflection tool to reassess both the produced output and the corresponding reasoning trajectory. This mechanism plays a pivotal role in the overall reasoning process. CoT reasoning 
primarily operates within the linguistic space, relying on symbolic inference without dynamic grounding to visual evidence, thereby limiting interpretability and robustness. By introducing the visual reflection module, the model can leverage explicit visual signals to cross-validate and iteratively refine its intermediate reasoning states, forming a closed-loop self-correction process that substantially improves the accuracy and consistency of final predictions.


\subsection{Visual Tools and Visual Reflection}\label{method:tool}
 
\begin{figure}[tbp]
  \centering 
  \includegraphics[width=0.9\linewidth]{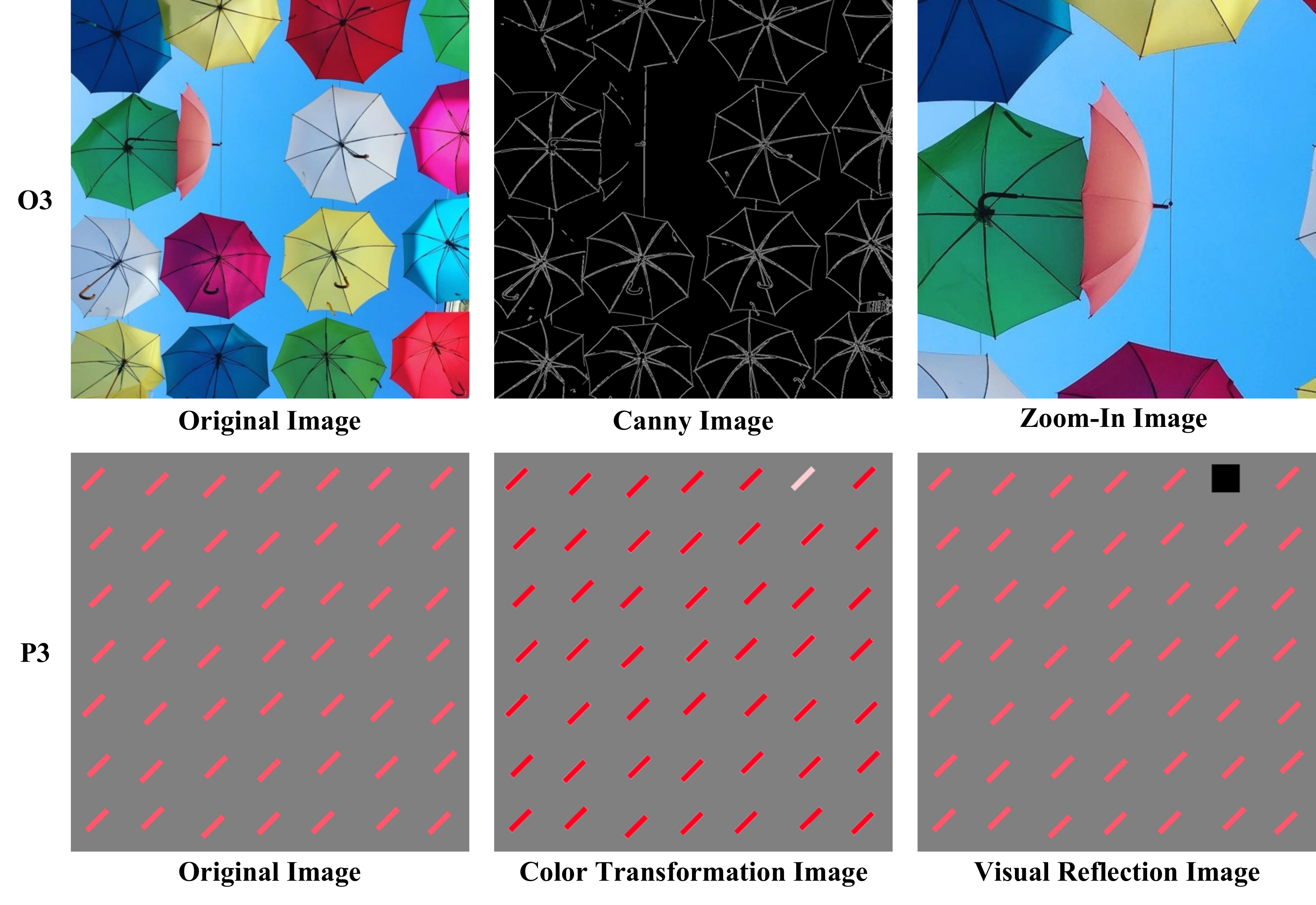}
  \caption{\textbf{Visualization of images using visual tools.} O3 represents natural image, and P3 represents synthetic data.} 
  \label{fig:vis_tool} 
\end{figure}

In this section, we provide a more detailed explanation of low-level visual tools and visual reflection in the proposed framework. Through the active invocation of these tools during the CoT reasoning process, the model achieves effective visual information parsing and reasoning enhancement. Fig.~\ref{fig:vis_tool} shows the visualization effects of visual tools and visual reflection.

\noindent
\textbf{Visual Tools.} ForeSight integrates a powerful visual toolkit, endowing it with the capability to acquire low-level visual information. These tools enable the model to examine images on demand during the reasoning process for critical information extraction. The toolkit comprises three core components: (1) \textbf{Canny Tool}. Built on the Canny algorithm for edge detection, this tool identifies object boundaries by detecting regions with strong intensity gradients. It provides explicit structural cues for the model, thereby facilitating contour-aware visual reasoning. (2) \textbf{Zoom-In Tool}. The visual zoom-in module allows the model to independently assess the sufficiency of visual details during CoT reasoning. When available information is inadequate, the model adaptively zooms in on relevant image regions to extract more fine-grained cues, supporting more accurate multimodal reasoning. (3) \textbf{Color Tool}. 
The color processing tool is introduced as a low-level visual component to amplify color differences in images, providing reliable cues that enhance the model’s grounding capability. Combined with CoT reasoning, the model can selectively invoke this tool to strengthen local image features, thereby improving the accuracy and robustness of subsequent reasoning.

\noindent
\textbf{Visual Reflection.} The visual reflection module is a core component of our ForeSight framework. Previous CoT-based models lack interactive reflection, and the thinking process is like a ``brain in a vat". Our visual reflection tool enables the model to validate and refine its reasoning iteratively, improving both accuracy and robustness.
Let $I$, $I_{\text{ans}}$, and $T_{\text{ans}}$ denote the input image, the grounding region predicted in the current reasoning step, and the corresponding predicted answer, respectively.
During the grounding reasoning process, the model records its output in \texttt{<draft\_answer>...</draft\_answer>}, containing both $T_{\text{ans}}$ and $I_{\text{ans}}$. After generating this draft answer, the visual reflection module is invoked to mask the previously grounded region and re-evaluate the reasoning. 

Specifically, we define $I'^{(k)}$ as the input image at the $k$-th iteration after masking the prior grounding region $I_{\text{ans}}^{(k-1)}$, and let $T_{\text{ans}}^{(k)}$ and $I_{\text{ans}}^{(k)}$ denote the updated answer and grounding region produced by the model at this iteration. Formally, this process is expressed as:

\begin{equation}
\begin{split}
I'^{(k)} &= I \odot \big( 1 - \mathbf{1}_{I_{\text{ans}}^{(k-1)}} \big), \\
T_{\text{ans}}^{(k)}, I_{\text{ans}}^{(k)} &= f_\theta \Bigl( I'^{(k)}, 
T_{\text{ans}}^{(k-1)}, I_{\text{ans}}^{(k-1)} \Bigr),
\end{split}
\label{eq:visual_reflection}
\end{equation}
\noindent where $\mathbf{1}_{I_{\text{ans}}^{(k-1)}}$ is the binary mask of the previously grounded region, $\odot$ denotes element-wise multiplication, and $f_\theta(\cdot)$ represents the model’s CoT reasoning and grounding function. If no new salient regions are detected in $I'^{(k)}$, the answer is considered correct. Otherwise, the model identifies potential errors and iteratively refines its prediction, thereby enabling self-verification and reasoning enhancement grounded in visual feedback. The maximum number of iterations in the experiment is 2.

\begin{figure}[tbp]
  \centering 
  \includegraphics[width=0.95\linewidth]{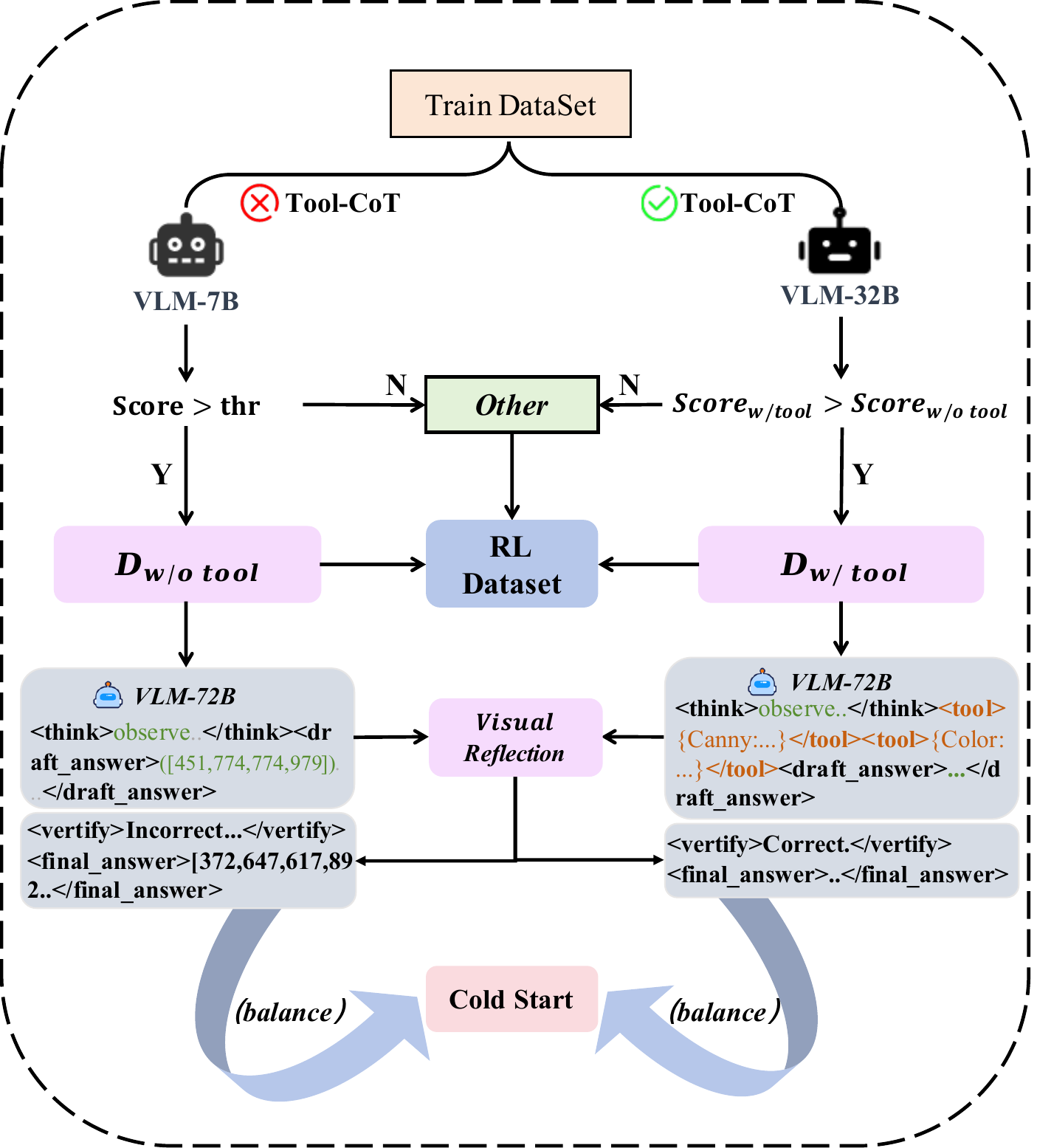}
    \caption{\textbf{V$^2$-CoT Data Construction Process.} All VLMs are based on the Qwen2.5-VL series. The raw dataset is split into the tool-free subset and the tool-dependent subset according to performance differences between 7B and 32B models. The 72B model generates structured CoT annotations with visual cues and reflection for both types, and rule-based cleaning yields a balanced, high-quality dataset for SFT and RL training. }
  \label{fig:data} 
\end{figure}

\begin{table*}[t]
\centering
\caption{Distribution of the 1425 Cold Start samples by tool usage pattern and source dataset. \texttt{ALL} means using all three tools.}
\label{tab:sft_dist}
\resizebox{0.75\textwidth}{!}{%
\begin{tabular}{@{}c|ccccccc|cc|c@{}}
\toprule
\begin{tabular}[c]{@{}c@{}}Data\\ Source\end{tabular}& \texttt{Canny} & \texttt{Zoom-in} & \texttt{Color} & \texttt{\begin{tabular}[c]{@{}c@{}}Canny\\ +Zoom-in\end{tabular}} & \texttt{\begin{tabular}[c]{@{}c@{}}Canny\\ +Color\end{tabular}} & \texttt{\begin{tabular}[c]{@{}c@{}}Zoom-in\\ +Color\end{tabular}} & \texttt{ALL} & \textbf{w/ tools} & \textbf{w/o tool} & \textbf{Total} \\ \midrule
O3          & 70             & 38           & 27             & 28                                                            & 44                                                              & 49                                                            & 57                                                                  & 313      & 185      & 498   \\
P3          & 104            & 155          & 39             & 80                                                            & 56                                                              & 51                                                            & 31                                                                  & 516      & 411      & 927   \\ \midrule
Total       & 174            & 193          & 66             & 108                                                           & 100                                                             & 100                                                           & 88                                                                  & 829      & 596      & 1425 \\ \bottomrule
\hline
\end{tabular}%
}
\end{table*}

\subsection{V$^2$-CoT Data Construction for Cold Start} \label{method:Training Data}

High-quality data needs to be constructed for supervised fine-tuning (SFT) before RL, serving as a cold start stage. To this end, we propose an automated data construction process for the \textbf{V}isual Cues and \textbf{V}isual Feedback chain (V$^2$-CoT). V$^2$-CoT data can provide the model with an initial framework for visual reasoning and visual reflection.
The entire process is illustrated in Fig.~\ref{fig:data}, encompassing three key steps: (1) Data Partitioning. Data where Qwen2.5-VL-7B achieves a high score without tool assistance is categorized as tool-free. In contrast, data where Qwen2.5-VL-32B performs better with tools than without them is thus categorized as tool-dependent. (2) CoT Annotation Generation. Qwen2.5-VL-72B is employed to generate CoT annotations for both data types, with visual reflection integrated into the CoT reasoning process. (3) Data Cleaning and Refinement. Manual rule-based data cleaning is conducted to filter out low-quality annotated samples, resulting in a balanced, high-quality SFT dataset. 

We implement the V$^2$-CoT construction procedure on the CG-SalBench training data. To support hybrid reasoning, balancing computational efficiency and reasoning reliability, we partition this data into distinct sets designated for SFT and RL. The resulting SFT dataset comprises 596 tool-free samples and 829 tool-dependent samples. This partition enables the model to learn when to rely on pure textual reasoning and when to invoke visual tools for perceptual enhancement during training. The detailed distribution across various tool usage patterns is presented in Table~\ref{tab:sft_dist}. This structured dataset enables the agent to simultaneously acquire efficient text-based reasoning and deliberate, perception-driven decision-making. More details are provided in the Appendix.

\subsection{Reinforcement Learning for V$^2$-CoT} \label{method:RL}
We employ a reinforcement learning paradigm to enable dynamic and autonomous tool utilization within the V$^2$-CoT framework. To effectively guide the model in generating reasoning trajectories that are well-structured, visually aligned, and semantically accurate during end-to-end RL, we design the following reward function.

\noindent \textbf{Format Consistency Reward ($R_{\text{fmt}}$)}~\cite{zhang2025r1}, which motivates the model to generate reasonable inference trajectories.

\noindent \textbf{IoU Reward ($R_{\text{IoU}}$)}~\cite{yu2016unitbox}, which uses the Intersection over Union (IoU) ratio to enhance the accuracy of localization.

\noindent \textbf{F1 Reward ($R_{\text{F1}}$)}~\cite{clark2016deep}, which improves recognition accuracy by calculating the F1 score between the generated text and the ground truth (GT).

\noindent \textbf{Tool Reward ($R_{\text{tc}}$).} To encourage correct tool usage, the tool reward function provides positive rewards exclusively when models answer questions accurately through proper tool usage. Formally, define $T_{\text{used}}$ as the tool usage indicator, $T_{\text{over}}$ as the overuse indicator, $F_{\text{invalid}}$ as the invalid format indicator, $A$ as the answer indicator value, and $\theta$ as the accuracy threshold. Except for $A$ and $\theta$, the other variables are binary indicators.  The tool reward function can be formulated as:
\begin{equation}
R_{\text{tc}}(\tau) = 
\begin{cases} 
1, & \text{if } T_{\text{used}} = 1 \land T_{\text{over}} = 0 \land A > \theta, \\
-1, & \text{if } T_{\text{over}} = 1 \lor F_{\text{invalid}} = 1, \\
0, & \text{otherwise}.
\end{cases}
\label{eq:tool_reward}
\end{equation}

\noindent \textbf{Verification Reward ($R_{\text{ver}}$).} To ensure the correctness of visual reflection, the verification reward function that aligns the verification result, draft answer, final answer, and GT discrepancy. Specifically, if the draft answer matches the GT, the semantic content within \texttt{<verify>...</verify>} must be correct and the final answer must match the draft. Conversely, if the draft deviates significantly from the GT, the \texttt{<verify>...</verify>} content must be incorrect and the \texttt{<final\_answer>...</final\_answer>} must be revised. Let $C_{\text{valid}}(\tau) \in \{0,1\}$ represent whether the trajectory $\tau$ satisfies the above condition. Then:
\begin{equation}
R_{\text{ver}}(\tau) = 
\begin{cases} 
1, & \text{if } C_{\text{valid}}(\tau) = 1, \\
-0.5, & \text{if } F_{\text{invalid}} = 1, \\
0, & \text{otherwise}.
\end{cases}
\label{eq:tool_reward}
\end{equation}
The total reward function $R(\tau)$ is evaluated on each complete reasoning trajectory $\tau$ as follows:
\begin{equation}
\begin{split}
R(\tau) & = \lambda_{\text{fmt}} \cdot R_{\text{fmt}}(\tau) + \lambda_{\text{IoU}} \cdot R_{\text{IoU}}(\tau) + \lambda_{\text{F1}} \cdot R_{\text{F1}}(\tau) \\
        &\quad + \lambda_{\text{tc}} \cdot R_{\text{tc}}(\tau) + \lambda_{\text{ver}} \cdot R_{\text{ver}}(\tau),
\end{split}
\label{eq:total_reward}
\end{equation}
where $\lambda_i$ are the weighting factors. In the experiment, the values are assigned as 0.2, 1.2, 1, 0.8, and 1.2, respectively. This ensures comprehensive supervision over the model's structural output and visual reasoning capabilities.

\begin{table}[tbp]
\centering
\small
\caption{Performance comparison of our method against open-source and closed-source SOTA models on the CG-SalBench benchmark. Best scores per column are highlighted in \textbf{bold}. ``+" indicates improved performance compared to the base model.}
\label{tab:main_results}
\resizebox{\linewidth}{!}{
\begin{NiceTabular}{lcccc}
\CodeBefore
  \rowcolor{gray!15}{2} 
  \rowcolor{gray!15}{7} 
\Body

\toprule
\textbf{Model} & \textbf{IoU} & \textbf{F1} & \textbf{Precision} & \textbf{Recall} \\
\midrule

\multicolumn{5}{c}{\textit{Open-Source Model}} \\[-0.25em]
\midrule
InternVL3-8B~\cite{zhu2025internvl3} & 30.50 & 72.09 & 69.74 & 79.01 \\
Qwen2.5-VL-7B~\cite{bai2025qwen2} & 32.56 & 64.97 & 61.46 & 75.81 \\
Qwen2.5-VL-72B~\cite{bai2025qwen2} & \textbf{66.00} & 82.70 & 82.37 & 86.74 \\
Qwen3-VL-8B~\cite{AlibabaCloud2025Qwen3VL8B} & 48.40 & 70.94 & 67.76 & 81.15 \\
\midrule

\multicolumn{5}{c}{\textit{Closed-Source Model}} \\[-0.25em]
\midrule
GPT-5~\cite{zhang2023complete} & 7.85 & \textbf{89.85} & 87.79 & \textbf{95.54} \\
Gemini-2.5-Flash~\cite{team2024gemini} & 16.41 & 88.82 & 87.32 & 93.91 \\
Doubao-Seed-1.6~\cite{ByteDance2024Seed} & 47.66 & 89.10 & \textbf{90.55} & 89.99 \\
Claude-Sonnet-4~\cite{Anthropic2025Claude4} & 20.82 & 82.50 & 82.40 & 89.13 \\
\midrule

\textbf{ForeSight (7B)} & \textbf{62.24} & \textbf{84.24} & \textbf{84.52} & \textbf{87.18} \\
\textbf{$\Delta$ (vs Qwen2.5-VL-7B)} & +29.68 & +19.27 & +23.06 & +11.37 \\
\bottomrule
\end{NiceTabular}
}
\end{table}

\section{Experiments}

\begin{table*}[htbp]
\centering
\small
\caption{\textbf{Ablation on visual tools and visual reflection.} Performance of closed-source models with tools invoked at inference time only (no training). Performance is presented for base, base with visual reflection (w/ reflect), and base with low-level visual tools (w/ tools). \textbf{Bold} values show improvement over base.}
\label{tab:inference_tool}
\resizebox{1\linewidth}{!}{
\setlength{\tabcolsep}{4pt} 
\begin{tabular}{l ccc ccc ccc ccc}
\toprule
Model & \multicolumn{3}{c}{IoU} & \multicolumn{3}{c}{F1} & \multicolumn{3}{c}{Precision} & \multicolumn{3}{c}{Recall} \\
\cmidrule(lr){2-4} \cmidrule(lr){5-7} \cmidrule(lr){8-10} \cmidrule(lr){11-13}
  & base & w/ reflect & w/ tools & base & w/ reflect & w/ tools & base & w/ reflect & w/ tools & base & w/ reflect & w/ tools \\
\midrule
GPT-5     & 7.85 & \textbf{14.92} & \textbf{12.54} & 89.85 & \textbf{89.96} & 87.75 & 87.79 & \textbf{89.04} & 84.36 & 95.54 & 93.28 & \textbf{96.42} \\
Gemini-2.5-Flash     & 16.41 & \textbf{18.12} & \textbf{17.70} & 88.82 & 67.83 & \textbf{91.10} & 87.32 & 59.83 & \textbf{90.26} & 93.91 & 92.95 & \textbf{95.05} \\
Doubao-Seed-1.6    & 47.66 & 46.04 & 34.73 & 89.10 & \textbf{89.24} & \textbf{91.54} & 90.55 & 90.63 & \textbf{93.46} & 89.99 & \textbf{90.18} & \textbf{91.54} \\
Claude-Sonnet-4     & 20.82 & \textbf{28.66} & \textbf{36.09} & 82.50 & 61.07 & 66.10 & 82.40 & 51.97 & 56.89 & 89.13 & \textbf{92.64} & \textbf{95.34} \\
\bottomrule
\end{tabular}}
\end{table*}

\subsection{Implementation Details}\label{method:implementation details}
The ForeSight model is implemented based on Qwen2.5-VL-7B and trained on 8 NVIDIA H20 GPUs via a two-stage pipeline: (1) SFT on 1425 samples with a global batch size of 32, a learning rate of $1 \times 10^{-4}$, for 10 epochs. (2) The RL stage uses async mode. 1 GPU is used for inference, and 7 GPUs are used for training. GRPO on 3929 samples with 7 rollouts per query, a global batch size of 28, and a learning rate of $1 \times 10^{-6}$ for 1 epoch. The sampling temperature is set to 0.9, the maximum number of calls for the Zoom\text{-}In tool is limited to 2, the maximum number of calls for other tools is limited to 1. 
For all evaluations, the sampling temperature is set to 0.7. All experiments use bfloat16 precision, a maximum sequence length of 12800 tokens, and a maximum pixel count of 1003520 per image. The experiment is implemented based on the SWIFT~\cite{zhao2024swiftascalablelightweightinfrastructure}.

\subsection{Evaluation Metrics}\label{method:evaluation metrics}

\noindent To comprehensively evaluate the model's performance on the CG-SalBench, we adopt four widely-used and essential metrics: IoU~\cite{lin2014microsoft}, the F1 score~\cite{christen2023review}, Precision~\cite{chen2024survey}, and Recall~\cite{singh2024benchmarking}. These metrics allow for a nuanced assessment of different aspects of the model's output. Specifically, IoU serves as the primary metric for quantifying the quality of Odd-One-Out Grounding by measuring the spatial overlap between the predicted and ground-truth bounding boxes. Conversely, Precision and Recall are utilized to assess the quality of Odd-One-Out Character Identification, reflecting the accuracy of positive predictions and the completeness of identification, respectively. Finally, the F1 score, defined as the harmonic mean of Precision and Recall, provides a single, balanced measure of identification performance.

\subsection{Main Results}\label{method:main results}

To comprehensively evaluate the proposed method, we conducted comparative experiments against a diverse set of state-of-the-art (SOTA) VLMs. Baselines include Open-Source models like InternVL3-8B~\cite{zhu2025internvl3}, Qwen2.5-VL-7B/72B~\cite{bai2025qwen2}, and Qwen3-VL-8B~\cite{AlibabaCloud2025Qwen3VL8B}, and leading Closed-Source API systems such as Doubao-Seed-1.6 ~\cite{ByteDance2024Seed}, Claude-Sonnet-4~\cite{Anthropic2025Claude4}, Gemini-2.5-Flash~\cite{team2024gemini}, and GPT-5~\cite{zhang2023complete}. Notably, ForeSight is built upon a 7B model, demonstrating superior performance via effective training, even at a modest scale. More visualizations are in the Appendix.

\begin{table}[ht]
\centering
\small
\caption{\textbf{Ablation on training components.} Evaluate different combinations of two training components: tools and reflect. The best result in each column is in \textbf{bold}.}
\resizebox{0.9\linewidth}{!}{
\label{tab:train_component}
\begin{tabular}{ccc cccc}
\toprule
tools & reflect & IoU & F1 & Precision & Recall \\
\midrule
\texttimes & \texttimes & 59.36 & 51.40 & 39.68 & 84.25 \\
\checkmark & \texttimes & 59.23 & 57.19 & 45.40 & 91.10 \\
\texttimes & \checkmark & 61.04 & 45.96 & 32.77 & \textbf{93.18} \\
\checkmark & \checkmark & \textbf{62.24} & \textbf{84.24} & \textbf{84.52} & 87.18 \\
\bottomrule
\end{tabular}
}
\end{table}

\begin{table}[ht]
\centering
\small
\caption{\textbf{Ablation on training stages.} Evaluate different combinations of SFT and RL. The best result in each column is in \textbf{bold}.}
\label{tab:training_paradigm}
\resizebox{0.85\linewidth}{!}{
\begin{tabular}{cc cccc}
\toprule
SFT & RL & IoU & F1 & Precision & Recall  \\
\midrule
\checkmark & \texttimes & 57.47 & 72.34 & 69.64 & 81.10 \\
\texttimes & \checkmark & 52.80 & 74.94 & 75.12 & 74.77 \\
\checkmark & \checkmark & \textbf{62.24} & \textbf{84.24} &  \textbf{84.52} & \textbf{87.18} \\
\bottomrule
\end{tabular}
}
\end{table}

\begin{table*}[htbp]
\centering
\caption{\textbf{Results on Grounding and VQA benchmarks.} We compare ForeSight with open-source MLLMs on several Grounding and VQA benchmarks. * denotes the results are reproduced by ourselves. \textbf{Bold} indicates an improvement compared to the baseline. }
\label{tab:generalization}
\resizebox{0.65\linewidth}{!}{
\begin{tabular}{l ccc cc} 
\toprule 
Model & \multicolumn{3}{c}{Grounding (Val)} & \multicolumn{2}{c}{VQA} \\
\cmidrule(r){2-4} \cmidrule(l){5-6} 
& RefCOCO & RefCOCO+ & RefCOCOg & MME & MMBench \\
\midrule
Qwen2.5-VL-72B & 92.7 & 88.9 & 89.9 & 2448 & 88.6 \\
Qwen2.5-VL-7B & 90.0 & 84.2 & 87.2 & 2347 & 83.5 \\
DeepEyes~\cite{zheng2025deepeyes} (7B) & 89.8 & 83.6 & 86.7 & - & - \\
Qwen2.5-VL-7B* & 88.91 & 82.11 & 86.4 & 2288.2 & 82.47 \\
\midrule
\textbf{ForeSight (7B)} & \textbf{91.42} & \textbf{83.66} & \textbf{88.43} & \textbf{2326} & 81.50 \\
\textbf{$\Delta$ (vs Qwen2.5-VL-7B*)} & +2.51 & +1.55 & +2.03 & +37.8 & -0.97 \\
\bottomrule
\end{tabular}
}
\end{table*}

As shown in Table~\ref{tab:main_results}, we present a comprehensive comparison against SOTA VLMs. Our 7B model achieves 62.24\% IoU and 84.24\% F1 score, positioning it among the top performers across all categories. Notably, our 7B model achieves an IoU close to that of the Qwen2.5-VL-72B (66.00\%) and significantly surpasses all other baselines in grounding, including the best closed-source model (Doubao-Seed-1.6, 47.66\%) by over 14\%. The Qwen2.5-VL-72B underperforms our method in F1 (82.70\% vs. 84.24\%), suggesting diminishing returns from scaling alone and validating our architectural choices. This superiority, despite the smaller scale, highlights the effectiveness of our low-level visual tools and reflection mechanism, demonstrating that our reasoning design successfully compensates for limited model size. Furthermore, while GPT-5 exhibites a high F1/Recall rate, its IoU is very low (7.85\%), indicating poor localization. Our method maintains high and balanced scores across all metrics, thus ensuring robust and accurate visual understanding. ForeSight enables human-like correction strategies, allowing the model to actively verify regions after an initial answer, which significantly mitigates the hard errors often caused by simple perceptual failures in current VLMs. This work demonstrates that equipping models with low-level, human-inspired perception tools is crucial for achieving robust, human-aligned visual intelligence, providing an alternative to relying solely on the model scaling law.

\subsection{Ablation Study}\label{method:ablation study}

\noindent \textbf{Ablation on Visual Tools and Visual Reflection.} We conduct a systematic ablation study to validate the design in our framework. First, we assess whether low-level visual tools and visual reflection can provide benefits without any specialized training. As shown in Table~\ref{tab:inference_tool}, we test several closed-source models under three inference settings: base, base with visual reflection (w/ reflect), and base with low\text{-}level visual tools (w/ tools). The results show that incorporating visual reflection and tools generally improves the IoU score for all models except Doubao-Seed-1.6, with a maximum gain of 15.27\%. Performance on Odd-One-Out Character identification also 
shows minor improvements, except for Claude-Sonnet-4. But the improvements are limited and inconsistent, confirming that simply providing tools during inference is insufficient. Models must be explicitly trained on when and how to effectively utilize them.

\noindent \textbf{Ablation on Training Components.} We ablate the impact of training with individual components subsequently. Table~\ref{tab:train_component} reports the results of our 7B model trained with different configurations. Using only low-level tools improves F1 from 51.40\% to 57.19\%, while using only visual reflection improves Recall from 84.25\% to 93.18\%. When both components are combined, performance surges across all metrics (IoU: 62.24\%, F1: 84.24\%). These results confirm their complementary roles: tools enable active perception, while visual reflection supports error correction.

\noindent \textbf{Ablation on Training Stages.} We further ablate the impact of our two-stage training strategy. Table~\ref{tab:training_paradigm} systemically validates our two-stage training paradigm. Compared to the SFT\text{-}only model (IoU 57.47\%, F1 72.34\%) and the unstable RL\text{-}only model (IoU 52.80\%), the full SFT + RL pipeline achieves the highest performance across all metrics (IoU 62.24\%, F1 84.24\%). This confirms that an initial cold start is crucial for establishing foundational skills, which are then effectively refined by reinforcement learning to achieve robust and optimal performance.

\subsection{Generalization Performance Analysis}\label{method:gpa}

To evaluate the generalizability of our proposed method beyond the primary task, we conduct experiments on three additional grounding benchmarks (RefCOCO~\cite{caesar2018coco}, RefCOCO+~\cite{caesar2018coco}, RefCOCOg~\cite{kazemzadeh2014referitgame}) and two Visual Question Answering (VQA) benchmarks (MME~\cite{fu2025mme}, MMBench~\cite{liu2024mmbench}). As shown in Table \ref{tab:generalization}, ForeSight is compared with several strong baseline models, including the Qwen2.5-VL model and DeepEyes. Our model achieves significant improvements across all grounding benchmarks, showing gains of +2.51\%, +1.55\%, and +2.03\%, respectively, and also leads on the MME VQA benchmark with an improvement of +37.8. Despite a slight decline on MMBench tests, the continued improvements in fundamental vision tasks and MME VQA tests validate the effectiveness of our training paradigm in enhancing fundamental vision and complex reasoning capabilities. ForeSight's improvement over DeepEyes on fundamental vision tasks is more significant. The findings further highlight the importance of dynamic and diverse low-level visual tools and reflection with visual feedback. More details of VQA are in the Appendix.

\section{Conclusion and Future Work}
In this paper, we propose ForeSight, an RL-driven multimodal reasoning framework. To address the insufficient low-level visual information and lack of effective visual reflection issues in previous VLMs, we incorporate sufficient low-level visual cues to enable the model to See Further and design an effective reflection with visual feedback to help the model Think Deeper. Through RL optimization, the model is empowered to autonomously determine whether to invoke tools and the order of tool invocation, rather than being confined to a fixed and rigid paradigm. Experimental results on the newly built dataset CG-Salbench show that ForeSight achieves better performance compared with the baseline counterpart. In the future, we will further extend this work to more visual tasks and validate its effectiveness.

{
    \small
    \bibliographystyle{ieeenat_fullname}
    \bibliography{main}
}

\clearpage
\setcounter{page}{1}
\maketitlesupplementary
\appendix
In this supplementary material, we provide further details on the dataset construction process and experimental results. Section \ref{sec:data-construct} introduces the dataset construction pipeline,  including prompt design and data template formulation used during data generation. Section \ref{sec:visual-cases} presents qualitative visualizations of model outputs, highlighting representative cases that further demonstrate the strengths and limitations of ForeSight. Section \ref{sec:vqa} provides detailed metrics on the model’s performance in general VQA tasks.
\section{Data Construction Pipeline}
\label{sec:data-construct}
\noindent
\textbf{Visual Tools Annotation.} We propose a novel framework for \textbf{V}isual Cues and \textbf{V}isual Feedback chain (V$^2$-CoT) data construction, designed to yield a high-quality, structured reasoning chain. 
The complex visual reasoning process is decoupled into discrete stages. Subsequently, manually configured rules are employed to integrate the data outputs from each stage, thereby assembling a comprehensive reasoning chain. This mechanism ultimately achieves a synergistic or closed-loop functionality encompassing ``thinking with visual cues" and ``reflection with visual feedback." Our approach utilizes a specific, structured prompt that guides the model toward tool-based reasoning, shown in \textit{Prompt 1}. This guidance enables the model to appropriately invoke designated visual tools, interpret the generated visual feedback, and seamlessly connect these intermediate steps to derive the final answer. This mechanism fundamentally realizes the principle of ``thinking with visual cues". The Canny tool, Zoom-In tool, and Color tool are referred to as $<$CANNY$>, <$ROI$>$, and $<$COLOR$>$, respectively. All tools are parameter-free, except for $<$ROI$>$.
For tool-free data instances,  \textit{Prompt 2} instructs the model to output image-based deep reasoning exclusively to derive a definitive result. This collaborative process, which integrates automated annotation with rule validation, leads to a substantial improvement in both the quality and the completeness of the generated visual reasoning data.

\noindent
\textbf{Visual Reflection Annotation.} The generation of ``Reflection with Visual Feedback" chains is achieved through a dedicated, automated pipeline. 
This process is mainly accomplished by \textit{Prompt 3}, which strategically guides the VLM to output a high-quality, efficient reflection chain.
The principle for the reflection stage is strictly conditional: if the predicted answer coincides with the ground-truth answer, the reflection process must produce a correct verification result based on the provided mask image. Conversely, in the event of a mismatch, the reflection mechanism is mandated to diagnose the error and propose a necessary correction. To ensure that the VLMs fully understand the reflection logic, our dataset specifically includes both the reasoning chains for reflecting correctly and the reasoning chains for reflecting incorrectly/proposing corrections.

\begin{figure}[htbp]
  \centering 
  \includegraphics[width=0.85\linewidth]{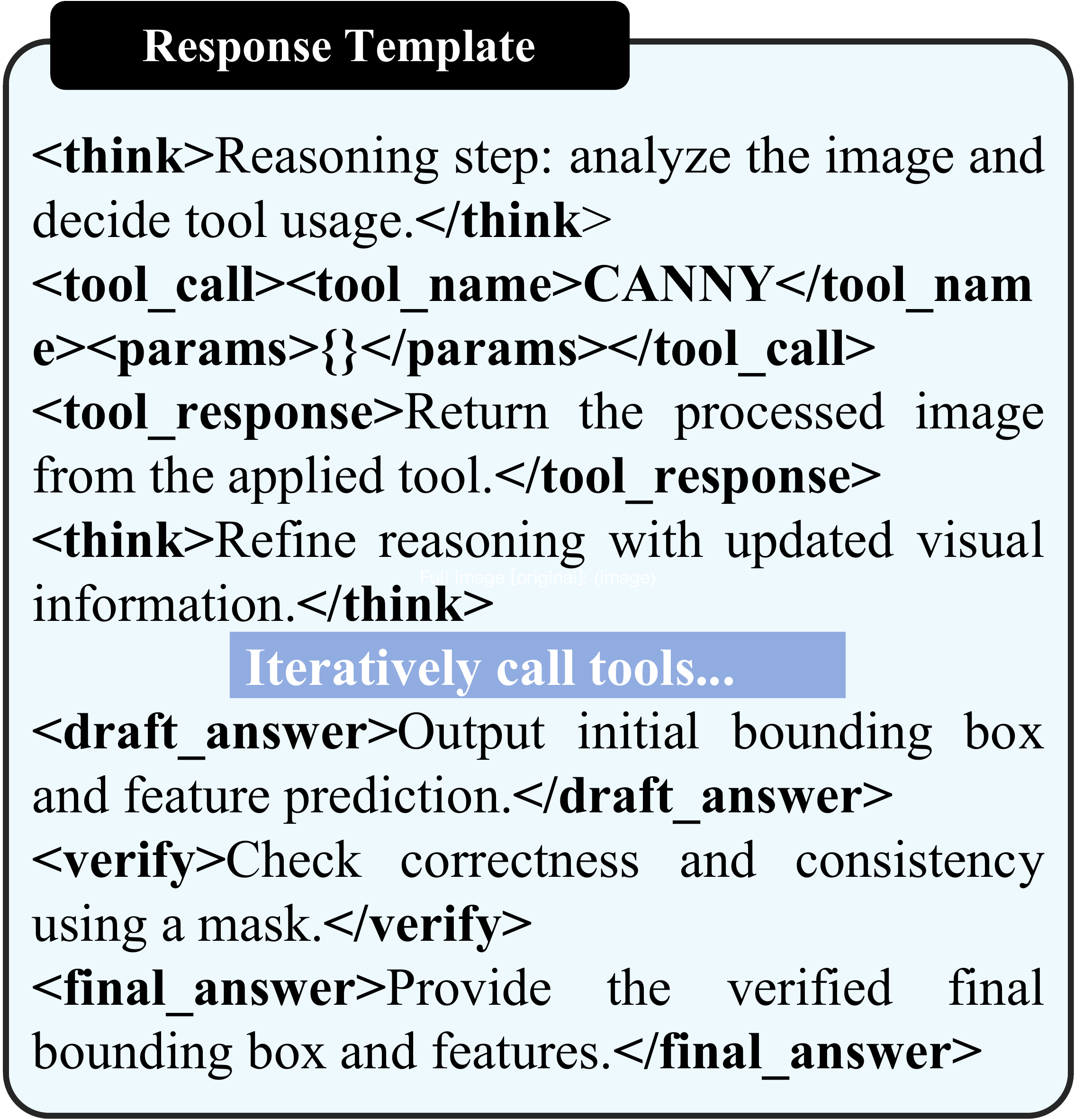}
  \caption{An example of data annotation.} 
  \label{fig:app_a} 
\end{figure}

\noindent
\textbf{Resulting Data Format.}
Following the two rounds of annotation (visual reasoning and visual reflection) the collected data is subjected to rigorous rule matching and cleaning. The final structured data annotations are then generated, as visually demonstrated in Fig.~\ref{fig:app_a}. The resulting dataset encapsulates a complete closed-loop reasoning. This process is initialized by a thought sequence that alternates between tool calls and the analysis of visual tool feedback, ultimately producing a draft answer. This draft answer subsequently serves as the basis for creating a mask on the original image, which provides the necessary visual feedback. Finally, the model leverages this generated visual feedback for in-depth reflection and validation, ultimately converging upon the final answer.

\begin{table}[t]
\caption{More details on VQA. FP is Fine-grained Perception.}
\label{tab:vqa}
\centering
\setlength{\tabcolsep}{2pt} 
\resizebox{0.5\textwidth}{!}{
\begin{tabular}{c|lcccc}
\hline
& \textbf{Metric} &
\begin{tabular}[c]{@{}c@{}}\textbf{Qwen2.5-VL}\\ \textbf{72B*}\end{tabular} &
\begin{tabular}[c]{@{}c@{}}\textbf{Qwen2.5-VL}\\ \textbf{7B*}\end{tabular} &
\begin{tabular}[c]{@{}c@{}}\textbf{ForeSight}\\ \textbf{7B}\end{tabular} &
\textbf{$\Delta$} \\ \hline

MME &
ALL & 2433.7 & 2288.2 & 2326 & +37.8 \\
&
Perception & 1702.9 & 1675.3 & 1704 & +28.7 \\
&
Reasoning & 730.7 & 612.9 & 622 & +9.1 \\ \hline

MMBench &
ALL & 89.18 & 82.47 & 81.50 & -0.97 \\
&
Attribute Reasoning & 87.94 & 78.89 & 78.68 & -0.21 \\
&
Coarse Perception & 89.86 & 88.18 & 86.47 & -1.71 \\
&
FP-Cross Instances & 90.91 & 74.83 & 75.50 & +0.67 \\
&
FP-Single Instances & 92.83 & 87.37 & 88.39 & +1.02 \\
&
Logical Reasoning & 80.51 & 72.88 & 71.97 & -0.91 \\
&
Relation Reasoning & 86.96 & 80.87 & 81.58 & +0.71 \\ \hline
\end{tabular}
}
\end{table}

\begin{figure*}
\begin{center}
\begin{tcolorbox}[colback=gray!5!white,colframe=green!60!black,width=0.95\linewidth,label={supp_prompt: iter}, title={\textit{\textbf{Prompt 1: Think With Tool Annotation Prompt }}}]
{
    {
\textbf{Character description}\\
You are an image analysis expert. You have the ability to analyze images, locate objects, and identify visual character.\\
\\
\textbf{Terminology Notes}:\\
-Target Character is a visual characteristic of an object in an image that is different from other objects.\\
-Target region is the coordinates of an object in an image that has different target characteristics from other objects.\\
\\
\textbf{Task Description}\\
-Using the provided auxiliary information, which includes the tools, target region, and target character, analyze and explain the reasons why the $<$Target Character$>$ in the image's $<$Target Region$>$ differs from the surrounding region. Explain why each tool from the auxiliary information was chosen and describe the results obtained.\\
-Use only the tools listed in the auxiliary information.\\
-Focus exclusively on the target region and feature.\\
-Provide a logical, coherent analysis following professional standards.\\
\\
\textbf{Analysis Steps}:\\
1. Initial Observation: Describe the general content of the image.\\
2. Tool Selection Reasoning: Explain why the first tool in the list was selected based on the initial observation.\\
3. Tool Application Results: Describe the results from using the first tool.\\
4. Subsequent Analysis: If multiple tools are listed, repeat steps 2 and 3 for each tool.\\
5. Conclusion: Based on the above findings, the target region and Target Character can be derived.\\
\\
\textbf{Output Requirements}:\\
1. Language: Use English.\\
2. Relevance: Ignore irrelevant information and avoid unnecessary details.\\
3. Format: Use natural and fluent language without timestamps or extra instructions.\\
4. Output Structure: Clearly separate each tool’s description and analysis sequentially.\\
5. Output Format: Use natural, fluent language. Provide a complete explanation without timestamps or additional instructions. Use `$<>$` to mark tools; choose from: $<$CANNY$>$, $<$ROI$>$, $<$COLOR$>$.
The sentence structure should be: "Based on the available information, we analyzed that we need to use $<$Tool 1$>$, and then use $<$Tool 2$>$ for further analysis.  $<$Attribution analysis$>$ shows that the chosen tools are sufficient to draw a conclusion."\\
6. Do not include bounding boxes $<$Target Region$>$ or Target Characters  $<$Target Character$>$ during the analysis; these are only shown in the final result.\\
7.  Descriptions of different tools should be separated and presented sequentially.\\
8. The tags $<$think$>$ and $<$/think$>$ cannot contain symbols such as $<$CANNY$>$, $<$ROI$>$, or $<$COLOR$>$.\\
9. Please output the result in the following format:\\
$<$think$><$Description, need to use tool xx$><$/think$><$tool xx$><$think$><$tool xx result description (whether other tools are used)$><$/think$>$(if other tools are used: $<$tool **$><$think$><$tool ** result description, final result$><$/think$>$)
\\
\\
\textbf{Example}:\\
• \textit{Example 1}\\

    }
}
\end{tcolorbox}
\end{center}
\end{figure*}

\begin{figure*}
\begin{center}
\begin{tcolorbox}[colback=gray!5!white,colframe=green!60!black,width=0.95\linewidth,label={supp_prompt: iter}, title={\textit{\textbf{Prompt 2: Think Without Tool Annotation Prompt }}}]
{
    {
\textbf{Character description}\\
You are an image analysis expert. You have the ability to analyze images, locate objects, and identify visual character.\\
\\
\textbf{Task Description}\\
Using the provided image and auxiliary information ($<$Target Region$>$, $<$Target Character$>$), analyze and explain the reasons why the $<$Target Character$>$ in the image's $<$Target Region$>$ differs from the surrounding region.
\\
\\
Note: Assume that the auxiliary information ($<$Target Region$>$, $<$Target Character$>$) is correct.
Do not output information about other regions or features. Your output should simulate the process of inferring the auxiliary information based solely on the given image, without prior knowledge of the $<$Target Region$>$ or $<$Target Character$>$.
Ensure your explanation is logically sound, well-organized, and adheres to professional analysis standards.
\\
\\
\textbf{Analysis Steps}:\\
Briefly describe the image content and explain the reasoning behind identifying the $<$Target Region$>$ and $<$Target Character$>$, ultimately leading to the analysis of these features.\\
\\
\textbf{Output Requirements}:\\
1. Language: Use English.\\
2. Filter irrelevant information: Ignore any irrelevant special characters and redundant information.\\
3. Coherence: Provide only explanations relevant to target identification; avoid unnecessary descriptions.\\
4. Output Format: Use natural and fluent language. Provide a complete explanation; no timestamps or other instructions are needed.\\
5. Do not include the target region bounding box $<$Target Region$>$ or target feature $<$Target Character$>$ in the analysis process.\\
6. Sentence structure: $<$description$>$, conducting $<$causal analysis$>$, we can identify the result.
Please output the result in the following format:
$<$think$><$Description, Analysis, Conclusion$><$/think$>$
\\
\\
\textbf{Example}:\\
• \textit{Example 1}\\

    }
}
\end{tcolorbox}
\end{center}
\end{figure*}

\begin{figure*}
\begin{center}
\begin{tcolorbox}[colback=gray!5!white,colframe=green!60!black,width=0.95\linewidth,label={supp_prompt_3}, title={\textit{\textbf{Prompt 3: Visual Reflection Annotation Prompt }}}]
{
    {
\textbf{Role Description:}\\
You are an image analysis expert capable of recognizing and describing image content based on visual character.\\
\\
\textbf{Task Description:}\\
-Use the available information (predictions) and auxiliary information (ground truth) to analyze and interpret whether the predicted answers in the available information are correct. By judging the masked part of the expected answer in the image, determine if the prediction is accurate, referencing the correct answer in the auxiliary information.\\
-If the predicted answer matches the ground truth, it is correct. In this case,  there should be no object of inconsistent visual character in the unmasked parts. The masked portion should represent a complete, single instance, and this instance should differ from the unmasked instances in the auxiliary information's features.\\
-If the predicted answer is different from the ground truth, the prediction result is wrong. In this case, there is a whole or part of the object with inconsistent visual character in the unmasked part.\\
\\
-Note: Auxiliary information is the ground truth. Do not output the auxiliary information. When outputting, you should pretend you only used the available information and did not know the auxiliary information in advance. Ensure the explanation is logical and clear, and that the final output meets professional analysis requirements.\\

\textbf{Analysis Steps:}\\
1. First, analyze whether there are targets with inconsistent visual character in the unmasked part of the image in the available information. visual character mainly include Orientation, Color, Focus, Shape, Size, Location, and Pattern.\\
2. Secondly, analyze whether the masked part of the image in the available information can completely cover the instance.\\
3. If both of the above points are true, then the prediction is correct, and the output is that the prediction is correct. Otherwise, the prediction is incorrect, and the output should suggest a corrected answer.\\
\\
\textbf{Output Requirements:}\\
1. Language: Use English.\\
2. Filter irrelevant information: Ignore any irrelevant special characters and redundant information.\\
3. Logical coherence: Provide only explanations relevant to object recognition; avoid unnecessary descriptions.\\
4. Output format: Use natural and fluent language. Provide a complete explanation; no timestamps or other instructions are needed.
Do not output terms like ``ground truth" or ``correct answer" during the analysis.\\
5. Sentence structure:$<$Description of visual character consistency$>,<$Description of the mask image$>,\\<$Analysis and conclusion$>$.\\
6. Please output the result in the following format:
$<$verify$><$Description, Analysis, Conclusion$><$/verify$>$ \\
\textbf{Example}:\\
• \textit{Example 1}\\
• \textit{Example 2}\\
    }
}
\end{tcolorbox}
\end{center}
\end{figure*}

\section{Qualitative Analysis}
\label{sec:visual-cases}
\begin{figure}[htbp]
  \centering 
  \includegraphics[width=\linewidth]{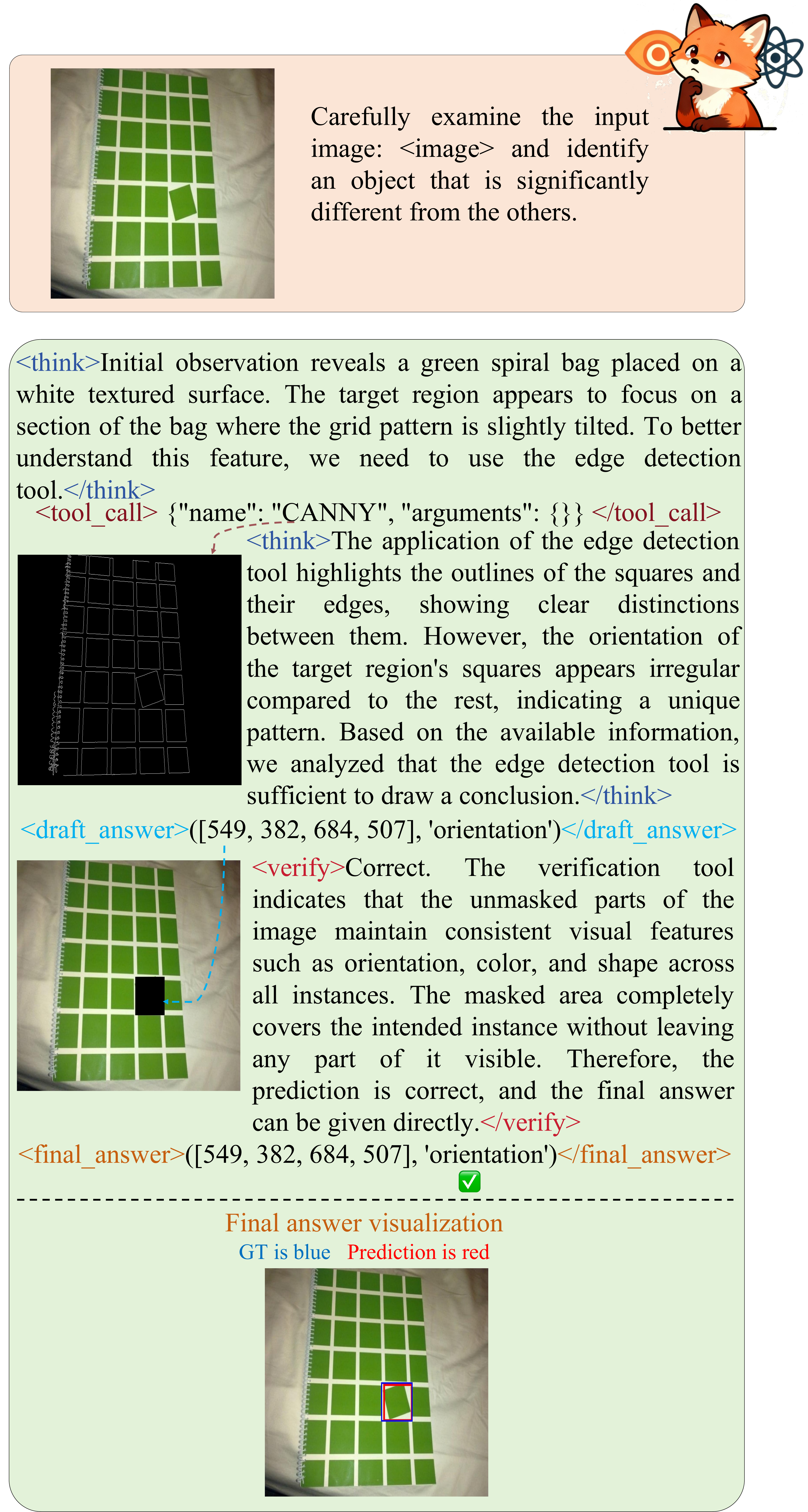}
  \caption{An example of using the Canny tool in natural image.} 
  \label{fig:o3can} 
\end{figure}

\begin{figure}[htbp]
  \centering 
  \includegraphics[width=\linewidth]{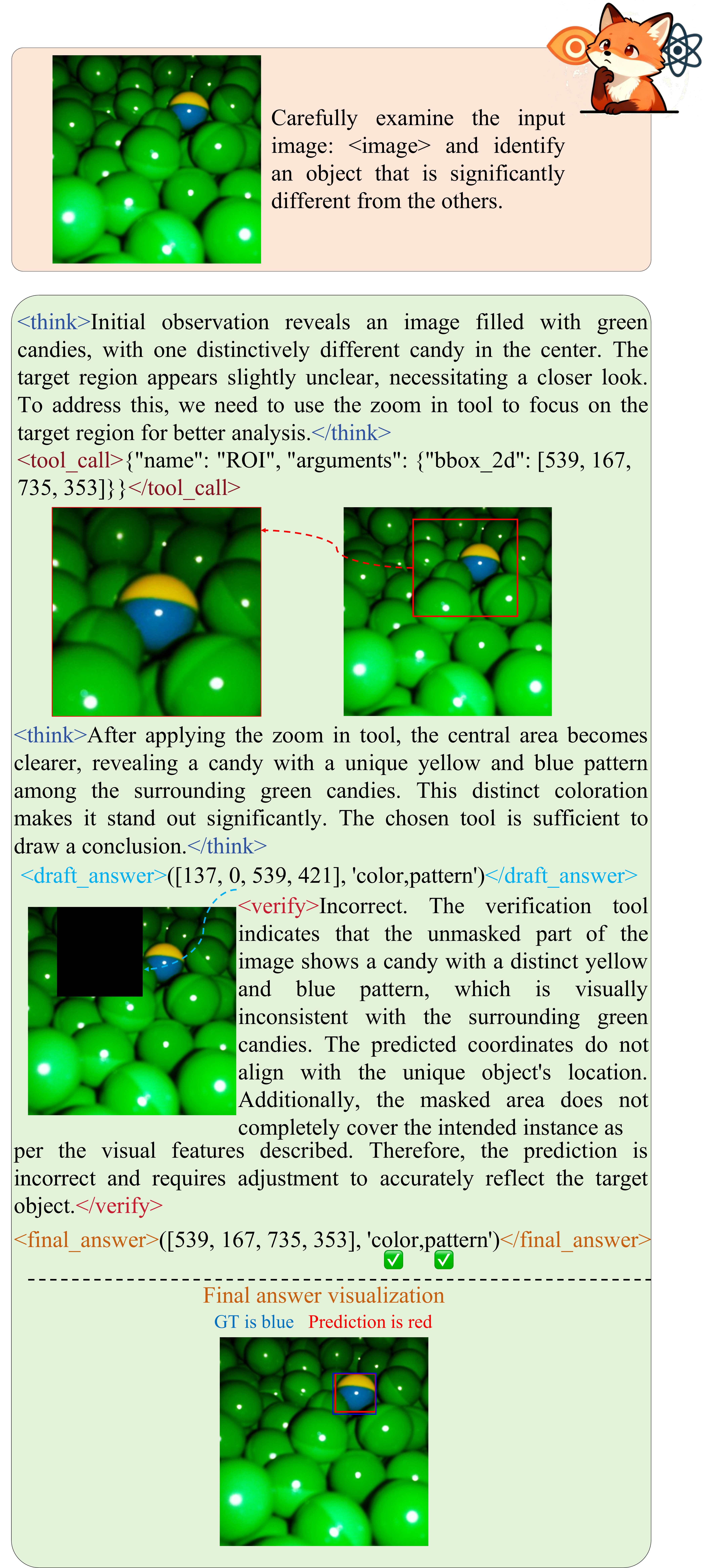}
  \caption{An example of using the Zoom-In tool in natural image.} 
  \label{fig:o3roi} 
\end{figure}

\begin{figure}[htbp]
  \centering 
  \includegraphics[width=0.85\linewidth]{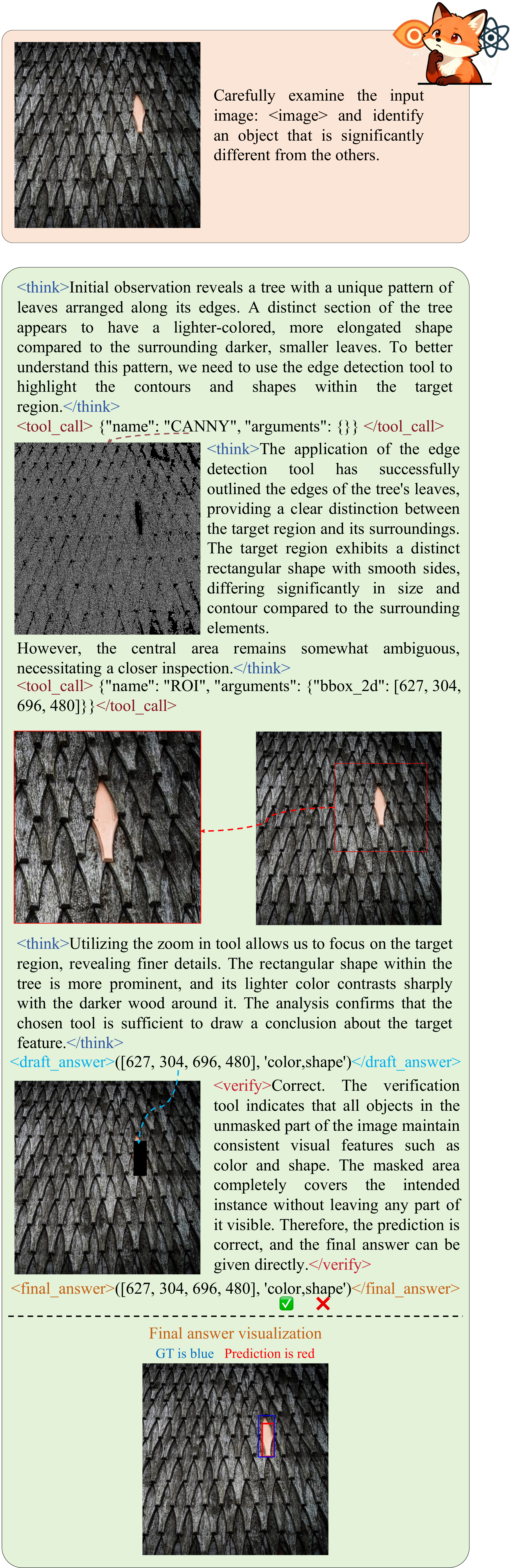}
  \caption{An example of using a combination of Canny and ROI tools in natural image.} 
  \label{fig:o3canroi} 
\end{figure}

\begin{figure}[htbp]
  \centering 
  \includegraphics[width=0.95\linewidth]{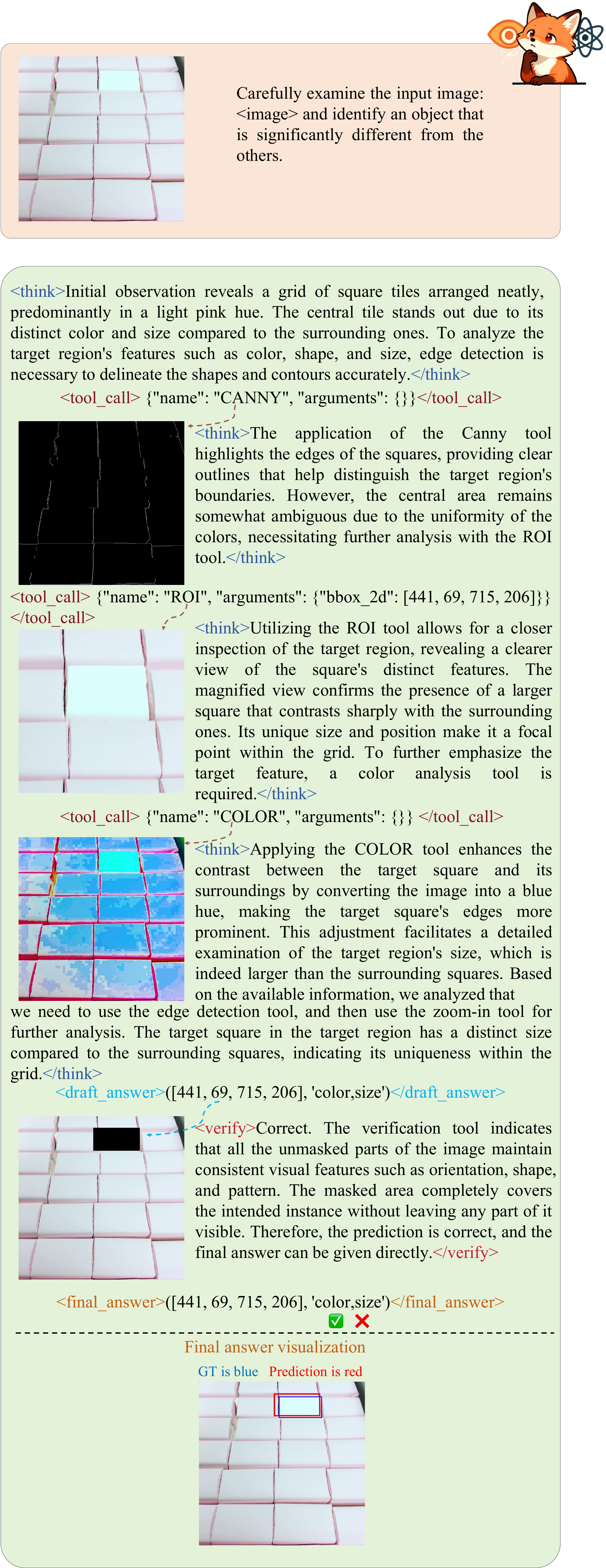}
  \caption{An example of using all tools in natural image.} 
  \label{fig:o3all} 
\end{figure}

\begin{figure}[htbp]
  \centering 
  \includegraphics[width=0.95\linewidth]{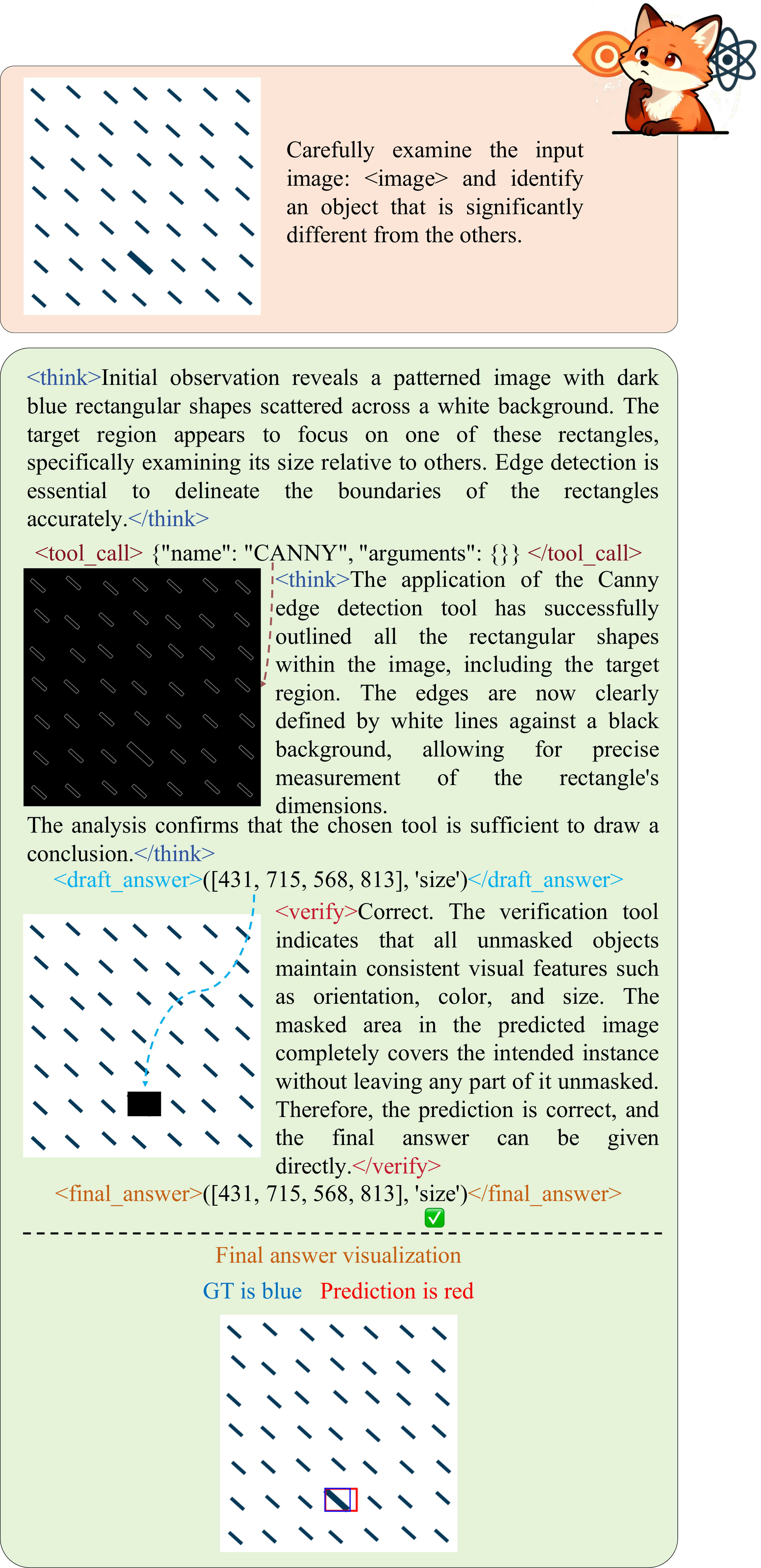}
  \caption{An example of using the Canny tool in synthetic image.} 
  \label{fig:p3can} 
\end{figure}

\begin{figure}[htbp]
  \centering 
  \includegraphics[width=\linewidth]{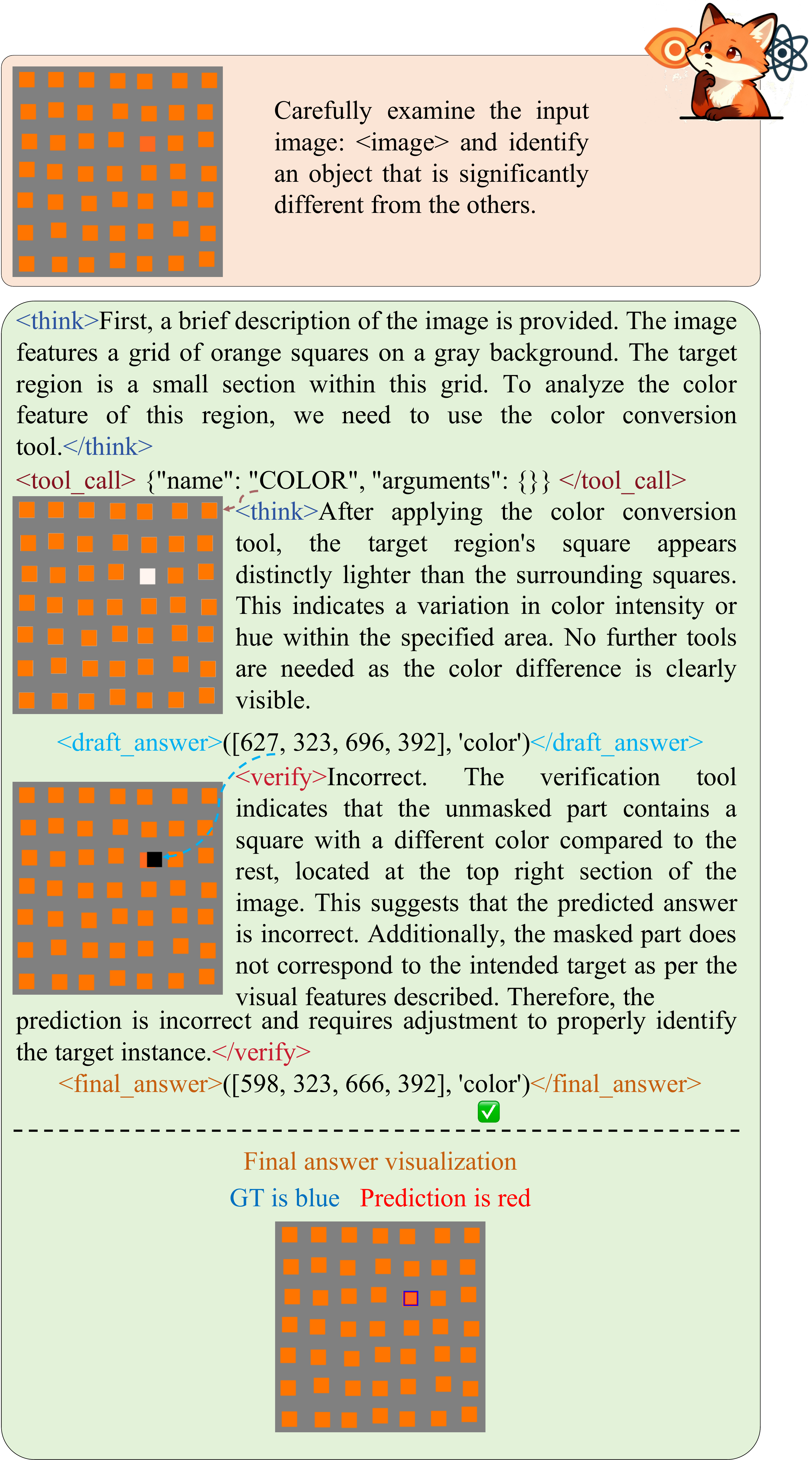}
  \caption{An example of using the Color tool in synthetic image.} 
  \label{fig:p3col} 
\end{figure}

\begin{figure}[htbp]
  \centering 
  \includegraphics[width=0.9\linewidth]{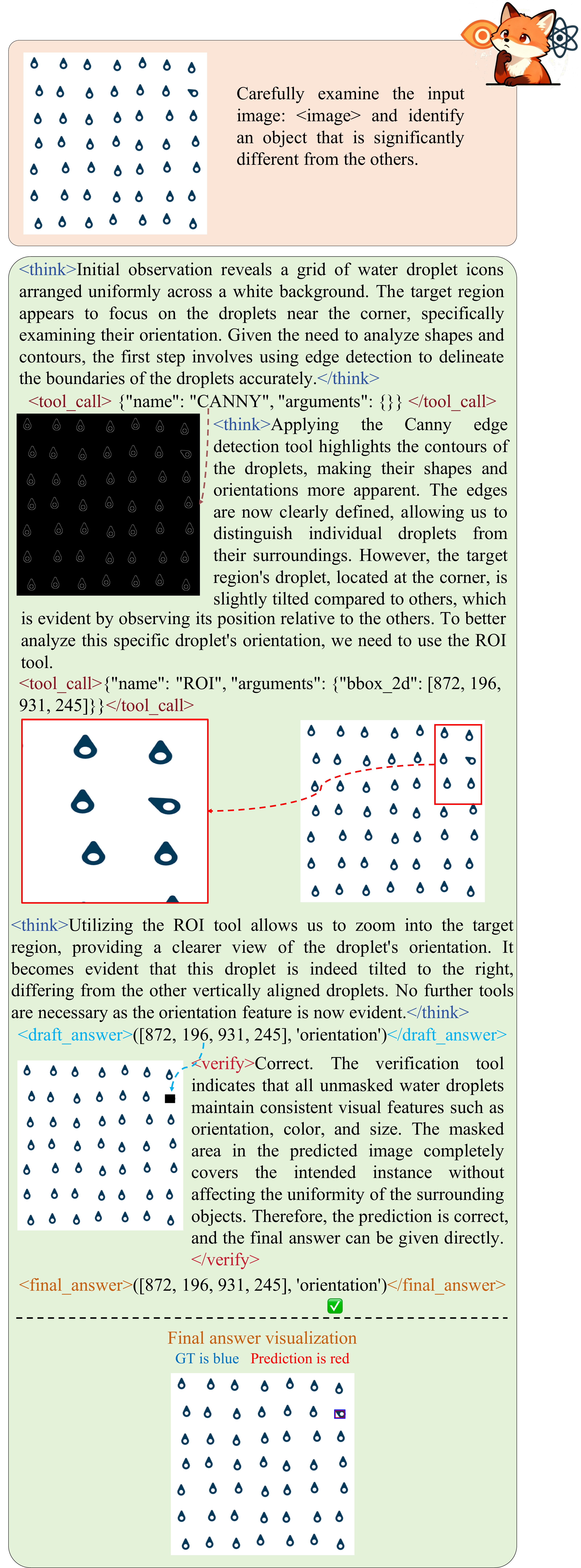}
  \caption{An example of using the Canny and ROI tool in synthetic image.} 
  \label{fig:p3canroi} 
\end{figure}

\noindent
\textbf{Cases of Visual Tools.}
This section provides qualitative examples to demonstrate the effectiveness of our proposed method. Fig.~\ref{fig:o3can}, Fig.~\ref{fig:o3roi}, Fig.~\ref{fig:o3canroi}, and Fig.~\ref{fig:o3all} illustrate the outcomes of various tool combinations applied to natural images, while Fig.~\ref{fig:p3can}, Fig.~\ref{fig:p3col}, and Fig.~\ref{fig:p3canroi} showcase these effects on synthetic images. Specifically, when the target region is perceptually ambiguous or difficult to discern, the Zoom-in tool is employed (Fig.~\ref{fig:o3roi}). To confirm subtle variations in the shape of the target object, the Canny tool provides crucial corroborating evidence (Fig.~\ref{fig:o3can}, Fig.~\ref{fig:p3can}). Furthermore, as illustrated in Fig.~\ref{fig:p3col}, challenging color differentiation in the original image is mitigated by color conversion, which renders objects of disparate colors clearly discernible, thus facilitating the model in reaching the correct conclusion. Finally, when a single visual tool proves insufficient to confirm a distinction, a strategic combination of multiple tools significantly aids the model's judgment by aggregating complementary visual feedback (Fig.~\ref{fig:o3canroi}, Fig.~\ref{fig:o3all}, Fig.~\ref{fig:p3canroi}).

\begin{figure}[htbp]
  \centering 
  \includegraphics[width=\linewidth]{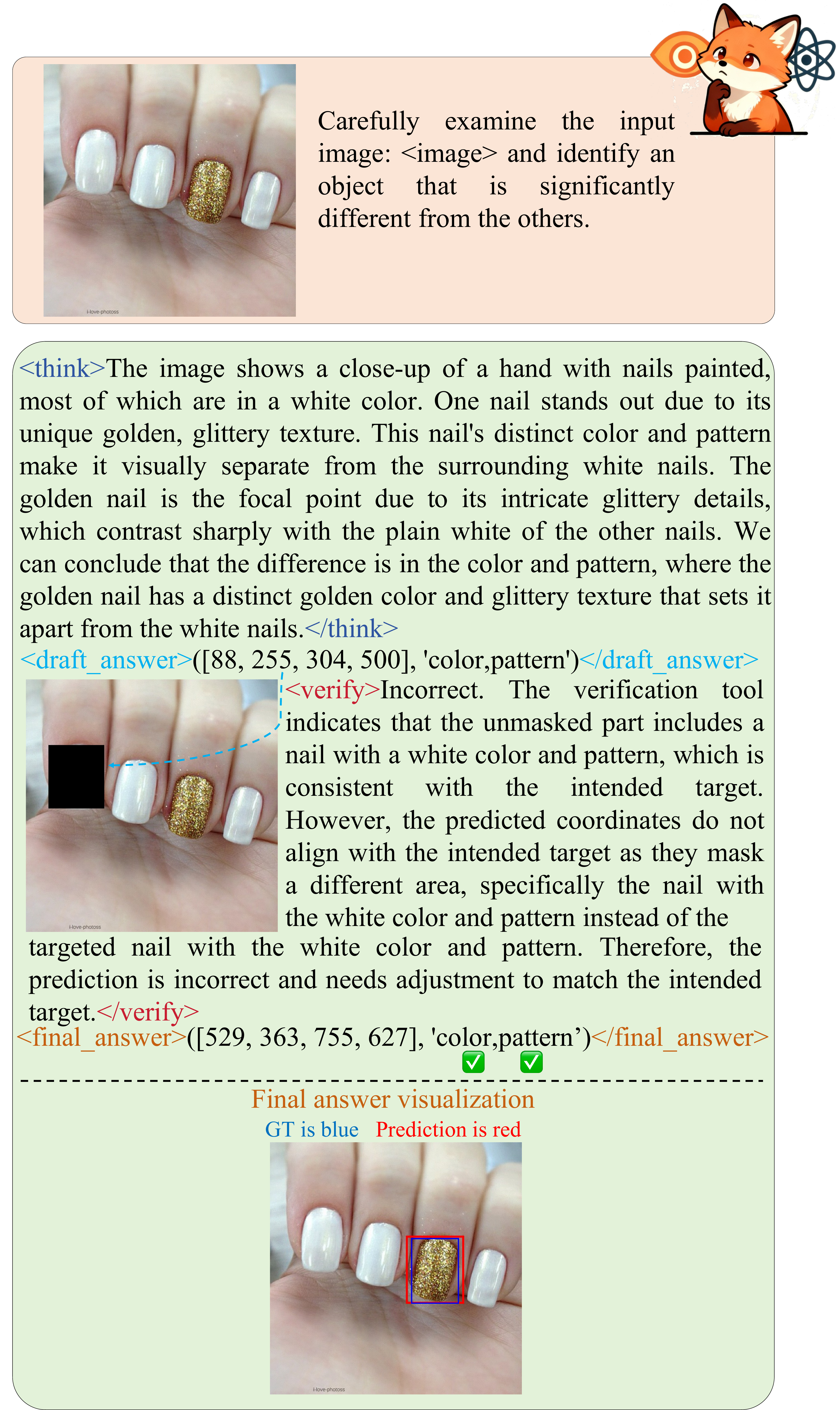}
  \caption{An example of visual reflection in natural image.} 
  \label{fig:vro3} 
\end{figure}

\begin{figure}[htbp]
  \centering 
  \includegraphics[width=\linewidth]{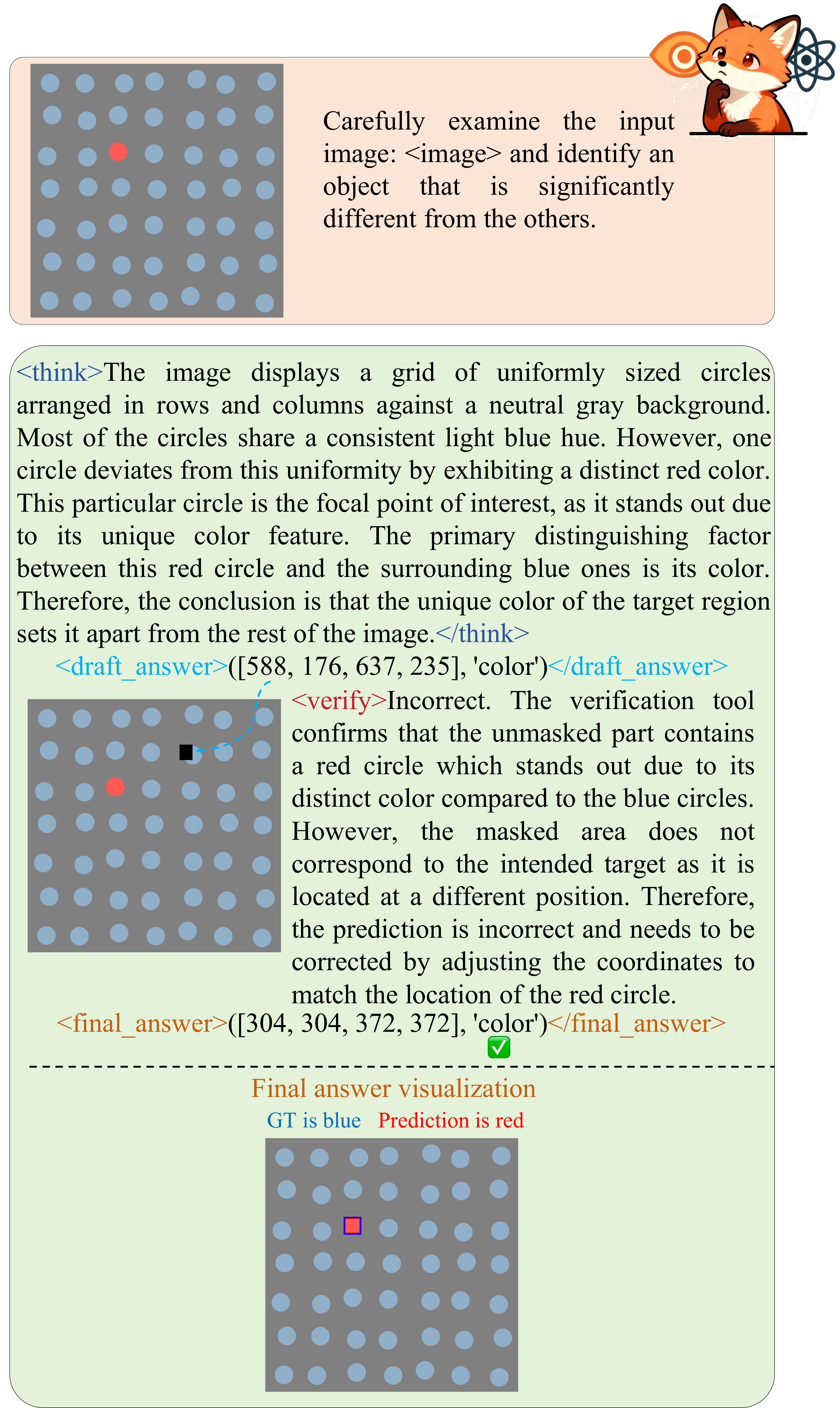}
  \caption{An example of visual reflection in synthetic image.} 
  \label{fig:vrp3} 
\end{figure}

\noindent
\textbf{Cases of Visual Reflection.}
Fig.~\ref{fig:vro3} and Fig.~\ref{fig:vrp3} demonstrate that ForeSight acquires the capability to execute visual reflection based on visual feedback during the inference phase, allowing it to successfully correct the incorrect draft answer and ultimately arrive at the correct conclusion.In particular, Fig.~\ref{fig:p3col} illustrates that the model retains the ability to identify errors even when the bounding box provided does not fully encapsulate the object's boundaries, thereby enabling the correction of the initial draft answer.

\noindent
\textbf{Failed Cases and Limitations.}
Fig.~\ref{fig:o3all} presents an example of a tool combination where the Canny tool appears to be redundant, which consequently contributed to an erroneous size assessment in the generated answer. Fig.~\ref{fig:o3canroi} shows a shape misjudgment, and crucially, the reflection module failed to detect the incomplete object encapsulation by the bounding box. These limitations are attributed partly to the relatively small size and weak foundational capabilities of the Qwen2.5-VL-7B model employed in this study. Future work will focus on adopting models with superior basic visual capabilities and expanding the coverage to encompass a broader spectrum of vision tasks.

\section{Details of VQA}
\label{sec:vqa}
Tab.~\ref{tab:vqa} provides more detailed VQA results. MME assesses perception (max 2000) and cognition (max 800), while MMBench evaluates 6 dimensions via iteratively questioning. ForeSight improves MME Reasoning by 9.1 and MMBench Relation Reasoning by 0.71, indicating that its benefits extend beyond basic perception grounding to visual reasoning. The gains mainly come from enhanced shape and positional accuracy, whereas the relatively modest overall improvement in VQA is likely due to the stronger reliance of some tasks on common sense and the performance drop in coarse perception caused by tool-induced changes in image style and quality.

\end{document}